\documentclass[sigconf]{acmart}

\newcommand{\wwwqy}{\color{black}}
\newcommand{\wwwqyz}{\color{black}}
\newcommand{\qywww}{\color{black}}
\newcommand{\qy}{\color{black}}
\newcommand{\zhu}{\color{black}}

\newcommand{\zlwww}{\color{black}}
\newcommand{\liang}{\color{black}}

\usepackage{graphicx}
\usepackage{enumitem}
\usepackage{subcaption}
\usepackage{multirow}

\renewcommand\footnotetextcopyrightpermission[1]{} 
\AtBeginDocument{%
  \providecommand\BibTeX{{%
    \normalfont B\kern-0.5em{\scshape i\kern-0.25em b}\kern-0.8em\TeX}}}

\begin{document}

\title{
HierPromptLM: A Pure PLM-based Framework for Representation Learning on Heterogeneous Text-rich Networks
}

\author{Qiuyu Zhu}
\affiliation{%
  \institution{Nanyang Technological University}
  \country{Singapore}
}
\email{qiuyu002@e.ntu.edu.sg}

\author{Liang Zhang}
\authornote{ Co-corresponding authors.}
\authornote{The work was done while Liang Zhang was a PhD student at NTU.}
\affiliation{%
  \institution{HEC Paris}
  \country{France}
}
\email{zhangl@hec.fr}

\author{Qianxiong Xu}
\affiliation{%
  \institution{Nanyang Technological University}
  \country{Singapore}
}
\email{qianxion001@e.ntu.edu.sg}

\author{Cheng Long}
\authornotemark[1]
\affiliation{%
  \institution{Nanyang Technological University}
  \country{Singapore}
}
\email{c.long@ntu.edu.sg}





\begin{abstract}

Representation learning on heterogeneous text-rich networks (HT- RNs), {\zlwww which consist} of multiple types of nodes and edges with each node associated with textual information, is essential for various {\zlwww real-world applications}. 
Given the success of pretrained language models (PLMs) in processing text data, recent efforts have focused on integrating PLMs into HTRN representation learning. These methods typically handle textual and structural information separately, using both PLMs and heterogeneous graph neural networks (HGNNs). However, this separation fails to capture the critical interactions between these two types of information within HTRNs. Additionally, it necessitates an extra alignment step, which is challenging due to the fundamental differences between distinct embedding spaces generated by PLMs and HGNNs. To deal with it, we propose \textbf{HierPromptLM}, a novel pure PLM-based framework that seamlessly models both text data and graph structures without the need for separate processing. Firstly, we develop a \emph{Hierarchical Prompt} module that employs prompt learning to integrate text data and heterogeneous graph structures at both the node and edge levels, within a unified textual space. Building upon this foundation, we further introduce two innovative HTRN-tailored pretraining tasks to fine-tune PLMs for representation learning by emphasizing the inherent heterogeneity and interactions between textual and structural information within HTRNs. Extensive experiments on two real-world HTRN datasets demonstrate HierPromptLM outperforms state-of-the-art methods, achieving significant improvements of up to 6.08\% for node classification and 10.84\% for link prediction.



\end{abstract}








\maketitle



\section{Introduction}
\label{sec:intro}

{\wwwqy Heterogeneous text-rich networks (HTRNs)~\cite{jin2023heterformer,shi2019discovering,zhang2019shne}, composed of multiple types of nodes and edges {\zlwww where} each node {\zlwww is} associated with textual information, have been widely used to model various {\qy web-world scenarios in real life, including social media~\cite{el2022twhin}, academic networks~\cite{tang2008arnetminer} and product networks~\cite{dong2020autoknow} in diverse web-based platforms.} 
For example, the academic data shown in Figure 1 can be represented as an HTRN, comprising three types of nodes (i.e., paper, author, venue) and three types of {\zlwww edges} (i.e., author-write-paper, venue-publish-paper, paper-cite-paper). {\zlwww Here, papers are associated with abundant textual information, including both titles and abstracts, which are commonly named text-rich nodes in previous works~\cite{jin2023heterformer,zou2023pretraining}. Venues, on the other hand, are associated only with name information, which are typically referred to as textless nodes~\cite{jin2023heterformer,zou2023pretraining}.} {\zlwww 
Due to their important roles, recent years have witnessed growing research interest in representation learning on HTRNs, which has emerged as a powerful tool for embedding nodes and/or edges in HTRNs into {\qywww meaningful} representations to facilitate various downstream tasks.}

\begin{figure}[th]    \centering\includegraphics[width=0.45\textwidth]{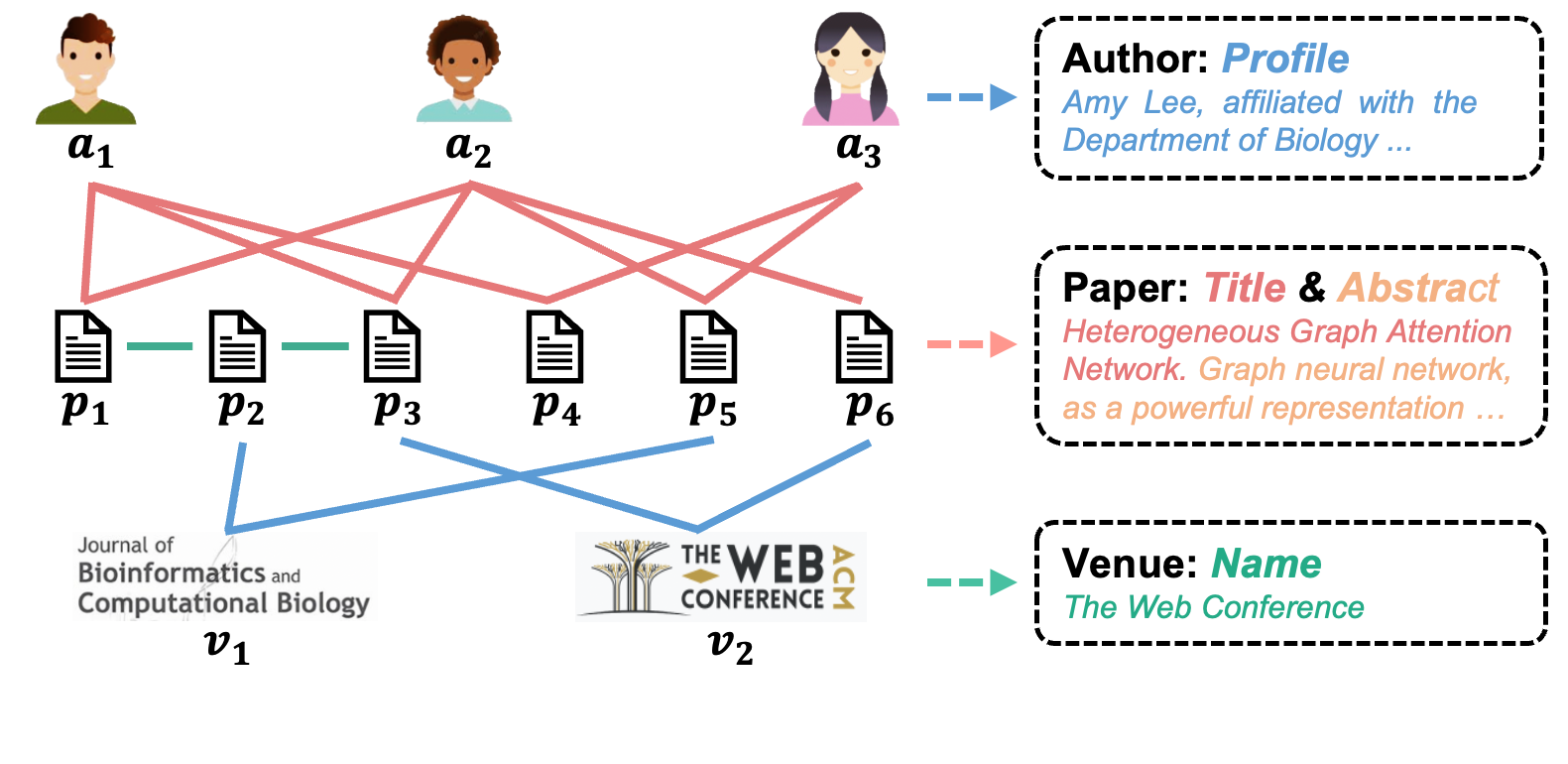}
\vspace{-6mm}
\caption{ An illustrative example of an HTRN.} 
\label{fig:intro}
\vspace{-3mm}
\end{figure}



{\zlwww The most intrinsic aspect of {\wwwqy HTRNs} is {\wwwqy their} heterogeneous structural information, which refers to the nodes {\qywww in HTRNs} and the heterogeneous connections among them. Heterogeneous Graph Neural Networks (HGNNs), which are models designed specifically for heterogeneous graph data, have been extensively utilized in existing studies 
{\qywww \cite{wang2019heterogeneous,fu2020magnn,yang2021interpretable,yang2022self}} 
to capture such structural information. For example,  HAN~\cite{wang2019heterogeneous} proposes a hierarchical neighbor aggregation strategy that adopts both node-level and meta-path-level attentions to effectively propagate and aggregate information across diverse
types of nodes in heterogeneous graphs. Building upon that, MAGNN~\cite{fu2020magnn} further considers intermediate nodes along meta-paths, leveraging both intra-meta-path and inter-meta-path information for high-order neighbor aggregations. However, these studies primarily focus on modeling the structural information, neglecting the additional textual information unique to {\wwwqy HTRNs}, which greatly limits their applicability to the HTRN representation learning problem.

To deal with the textual information in HTRNs, recent models have further extended HGNNs by integrating additional pretrained language models (PLMs)~\cite{kenton2019bert,liu2019roberta,chi2021audio} component, drawing inspiration from the significant success of PLMs in natural language processing {\qywww  (NLP) ~\cite{kenton2019bert,chi2021audio,radford2019language,2020t5}}. For example, Heterformer \cite{jin2023heterformer} learns a target node's representation in HTRNs by aligning the structural embedding generated by HGNNs from its one-hop neighbors with the textual embedding produced by auxiliary PLMs. 
THLM \cite{zou2023pretraining} also leverages PLMs to enrich node representations on HTRNs, making them text-aware by aligning the embedding spaces of HGNNs and PLMs through a binary matching task. {\liang Despite recent improvements, current methods process structural and textual information separately, overlooking the crucial interactions between them within HTRNs.
{\qy Specifically, consider a typical node classification scenario for an author node, illustrated in Figure~\ref{fig:intro}, where author $a_3$ frequently collaborates with other researchers in machine learning area (i.e., $a_1$ and $a_2$), while $a_3$'s own textual profiling indicates a primary focus on biology.} By explicitly integrating the interplay between textual and structural information, it becomes clear that the author applies machine learning techniques to biological problems, correctly identifying $a_3$ as working in computational biology field. Additionally, existing approaches that rely on separate processing often necessitate extra alignment, which is challenging due to the fundamental differences in the embedding spaces generated by PLMs and HGNNs. {\qy For instance, in the HTRN depicted in Figure~\ref{fig:intro}, the biology paper $p_2$ has several machine learning papers (i.e., $p_1$ and $p_3$) as neighbors resulting from citation relations. Enforcing alignment between this paper's textual embedding,} which indicates a focus on biology, and its structural embedding, which aggregates information from machine learning-focused neighbors, may introduce misleading signals to the model learning. 

To overcome these shortcomings associated with separate processing, we adopt a fundamentally different perspective and propose to jointly {\qy incorporate}
both textual and structural information within HTRNs in a unified representation space for the first time. To achieve this, we introduce a pure PLM-based framework, named \textbf{HierPromptLM} (\textbf{Hier}archical \textbf{Prompt} \textbf{L}anguage \textbf{M}odel),  given that PLMs have learned massive context-aware knowledge and demonstrated promising results in encoding {\qy either  text~\cite{kenton2019bert,2020t5,liu2019roberta} or graph data~\cite{ye2024language,fatemi2023talk,guo2023gpt4graph}.}
Our framework features a specialized \emph{Hierarchical Prompt} module, which leverages prompt learning to capture heterogeneous graph structures in HTRNs, seamlessly integrating them with textual information into a unified textual space, where interactions can be captured naturally and without the need for extra alignment. Unlike existing methods~\cite{ye2024language,guo2023gpt4graph,tang2024graphgpt} that primarily focus on homogeneous graphs, \emph{Hierarchical Prompt} introduces an automated textualization mechanism specifically designed for heterogeneous graphs and {\qy emphasizes} the inherent heterogeneity within HTRNs at both node and edge levels, which leads to two corresponding designs: the graph-aware prompt and the relation-aware prompt. We explain their intuitions and key ideas as follows. 
{\qy \textbf{(1) Graph-aware prompt} extracts and composes meaningful text sequences from the subgraph context surrounding a target node, enriching the node’s representation from a structural perspective in addition to its own textual information. Due to the inherent heterogeneity in a node's subgraph context, we decompose it into distinct meta-path-based subgraphs using pre-defined meta-paths, as shown in Figure~\ref{fig:model}(a), inspired by their effectiveness in capturing heterogeneous graph structures \cite{wang2019heterogeneous,hu2020heterogeneous,fu2020magnn}. }
These subgraphs enhance the target node's representation through structural augmentation, enabling to capture a broader context while distinguishing the fine-grained semantics of different neighbors through various meta-paths. By designing a meta-path-based subgraph program function to textualize these subgraphs into corresponding graph summaries and combining them with the node's own text data, we create a unified textual space to jointly handle both types of information. {\qy However, {\liang such a straightforward approach often generates} lengthy inputs, making it difficult for PLMs to process due to token length limitations~\cite{bi2024lpnl}.}
To address this issue, we employ a frozen PLM to distill the subgraph summaries into graph tokens before concatenation, which serve as specialized soft prompts and result in a hierarchical graph-aware prompt for each target node.
{\qy \textbf{(2) Relation-aware prompt} further captures the heterogeneous edge-level structural information within HTRNs.}
To explicitly capture the heterogeneity of relations, we introduce a learnable relation token, another special soft prompt designed to capture diverse interactions between different node pairs. This token is positioned between the graph-aware prompts of a pair of nodes, forming the final relation-aware prompt. Through such prompt design, we reformulate both the heterogeneous structural and textual information unique to HTRNs into a unified text sequence, marking the first attempt in this field to the best of our knowledge.

The obtained relation-aware prompt serves as input to a tunable PLM for representation learning. Instead of simple graph reconstruction \cite{jin2023heterformer} or binary matching \cite{zou2023pretraining}, we introduce two newly designed pretraining tasks specifically tailored for HTRNs in this paper, {\qy named \textbf{H}eterogeneous \textbf{G}raph-\textbf{a}ware \textbf{M}asked \textbf{L}anguage \textbf{M}odeling (HGA-MLM) and \textbf{H}eterogeneous \textbf{G}raph-\textbf{a}ware \textbf{N}ext \textbf{S}entence \textbf{P}rediction (HGA-NSP)}, to help emphasize the heterogeneity and text-rich characteristics inherent in HTRNs. Once trained, our framework can produce versatile, unsupervised representations for both nodes and edges in HTRNs, which can be utilized to facilitate various downstream applications. The main contributions of our paper are summarized as follows:

}

\begin{itemize}[leftmargin=*]
    \item For the first time, we propose a novel, pure PLM-based method, HierPromptLM, for representation learning on HTRNs, which features in an innovative hierarchical prompt framework that seamlessly integrates textual information with heterogeneous node-level and edge-level structures, along with their interactions, into a unified representation space.

    \item To the best of our knowledge, we are the first to propose two novel HTRN-tailored pretraining tasks for PLMs, enhancing representation learning on HTRNs.

    \item Experiments on two real-world HTRN benchmark datasets demonstrate that our model significantly outperforms 9 baselines on two typical downstream tasks, achieving notable improvements of up to 6.08\% for node classification and 10.84\%
    for link prediction on the DBLP dataset compared to state-of-the-art methods.
    
\end{itemize}


}

\section{Related Work}
\label{sec:related}

\subsection{Pretrained Language Models}
{\wwwqyz The primary goal of PLMs ~\cite{2020t5,kenton2019bert,chen2024llaga,openai2023gpt} 
is to learn general text representations from large-scale, unlabeled corpora through pretraining tasks, which can be applied to various downstream NLP tasks~\cite{liu2019fine,meng2022topic,xun2020correlation}. Early developments in PLMs, such as word2vec \cite{mikolov2013distributed} and GloVe~\cite{pennington2014glove}, focus on context-free embeddings. Driven by the fact that a word’s meaning can vary depending on its context, models such as BERT~\cite{kenton2019bert}, Alberta~\cite{chi2021audio},  
DistilBERT~\cite{sanh2019distilbert},
T5~\cite{2020t5}
and GPT~\cite{brown2020language,radford2019language}, are introduced to generate contextualized token representations that adapt to their surrounding text. However, these models mainly focus on text encoding {\zlwww  only. Recently, several studies~\cite{yasunaga2022linkbert,levine2021inductive,chien2021node} have been proposed to further enhance the capture of connections between different text.}
For instance, LinkBERT~\cite{yasunaga2022linkbert}, introduces a document relation prediction task that pretrains language models by classifying relations between text segments based on document links. GIANT~\cite{chien2021node} incorporates graph topology into language models by predicting neighbors that are connected to each target node. Despite these advancements, none of them are specifically designed for HTRNs, which requires to consider both textual information and the heterogeneity of relations. To address this gap, we propose two newly HTRN-tailored pretraining tasks, aiming to jointly capture the inherent heterogeneity and text-rich characteristics of HTRNs.}


\vspace{-2mm}
\subsection{Text-rich Network Mining}
Text-rich networks (TRNs){\wwwqyz ~\cite{jin2021bite,jin2023patton,yu2023teko,yu2021gcn,yu2023embedding}}, where nodes contain textual information, are widely used in real-world applications. With the success of PLMs~\cite{kenton2019bert,radford2019language} in handling text data, existing methods~\cite{yang2021graphformers,zhu2021textgnn,li2021adsgnn} {\zlwww propose to combine} both GNNs and PLMs to explore the textual and structural information within TRNs. However, these methods often assume networks are homogeneous, which is impractical in real life. Therefore, recent research has shifted toward representation learning on HTRNs~\cite{jin2023heterformer,zou2023pretraining}. For example, Heterformer~\cite{jin2023heterformer} is developed to embed nodes by combining structural embeddings from its direct neighbors generated by HGNNs with textual embeddings from auxiliary PLMs. To capture higher-order structural information, THLM~\cite{zou2023pretraining} aligns the embedding spaces of HGNNs and PLMs through a context graph prediction task.
{\qy While these methods have shown improvements, they treat structural and textual information separately, failing to fully capture the interactions between these two types of information. Moreover, this {\zhu separation typically requires an additional alignment step}, which poses a challenge due to the inherent differences between the embedding spaces encoded by PLMs and HGNNs.} Different from these works, {\qy we adopt a fundamentally different perspective by proposing the first pure PLM-based framework that jointly models both textual and heterogeneous structural information within HTRNs in a unified representation space.} 

\section{Preliminaries}
\label{sec:pre}

\noindent \textbf{Definition 3.1. Heterogeneous Text-Rich Networks.}
A heterogeneous text-rich network (HTRN) is defined as $\mathcal{G} = \{\mathcal{V}, \mathcal{E}$, $\mathcal{C}, \mathcal{R}, \mathcal{T}\}$, where $\mathcal{V}, \mathcal{E}, \mathcal{C}, \mathcal{R}, \mathcal{T}$ represent the set of nodes, edges, node types, relation types and textual descriptions, respectively. Each node is associated with a {\zlwww node} type $c\in \mathcal{C}$ and the textual information $t \in \mathcal{T}$, and each edge {\zlwww is characterized by a relation type} $r\in \mathcal{R}$. 
Note that an HTRN should satisfy the condition $|\mathcal{C}| + |\mathcal{R}| > 2$.

\noindent \textbf{Definition 3.2. HTRN Representation Learning Problem}.
Given an HTRN, our goal is to fine-tune a PLM $f_{\Theta}$ {\zlwww parameterized by ${\Theta}$, which effectively captures the textual and heterogeneous structural information of the HTRN and can be utilized to generate both node and edge representations to facilitate various downstream tasks.} Specifically, for a node $u$ in the HTRN, we aim to learn its embedding {\zlwww $z_u = f_{\Theta}(u)$, where $z_u \in \mathbb{R}^d$, and $d$ denotes the embedding dimension.} 
{\zhu Similarly, given a pair of nodes $u$ and $v$ connected by relation $r$, which forms an edge $e=(u,r,v)$,}
we aim to learn the edge embedding {\zlwww $z_{e} = f_{\Theta}(e)$ where $z_{e} \in \mathbb{R}^d$.}

\section{Methodology}
\label{sec:methodology}


In this section, we {\zlwww introduce} {\qywww our} method \textbf{HierPromptLM}, with its overall architecture depicted in Figure~\ref{fig:model}. {\wwwqy This method consists of two important components: (1) \emph{Hierarchical Prompt} first creates a graph-aware prompt by augmenting the target node's representation with its surrounding subgraph context to capture {\zhu the} heterogeneous node-level {\zlwww structural} information, followed by a relation-aware prompt to further capture the edge-level heterogeneity from HTRNs. Specifically, for a target node, meta-path-based subgraphs are first extracted to provide local structural context, which are then distilled into graph tokens using a frozen PLM. These graph tokens, along with the node's own textual information, are then combined via a graph-aware program function to construct the graph-aware prompt. To explicitly model the heterogeneity of {\zlwww edges in HTRNs}, a learnable relation token is {\zhu further} introduced, positioned between the graph-aware prompts of node pairs, forming the final relation-aware prompt. By employing {\zhu such} prompt design, we reformulate both textual and heterogeneous structural information unique to HTRNs into a unified text sequence. This approach provides a coherent textual representation that is easily interpretable by PLMs, facilitating the capture of {\zhu critical} interactions and eliminating the need for an additional alignment.
{\qy (2) \emph{HTRN-tailored Pretraining} introduces two novel tasks, HGA-NSP and HGA-MLM, for fine-tuning PLMs with relation-aware prompts as inputs. Specifically, HGA-NSP predicts whether two nodes are connected by a specific relation, {\zhu modeling heterogeneous relations in an explicit manner.} Meanwhile, HGA-MLM predicts masked tokens from both text and graph tokens in the prompt, capturing {\zhu crucial} interactions between text {\zhu data} and structures, thereby highlighting the heterogeneity and {\zlwww rich textual context inherent in HTRNs.}}
}

\begin{figure*}[th]
    \centering     \includegraphics[width=0.88\textwidth]{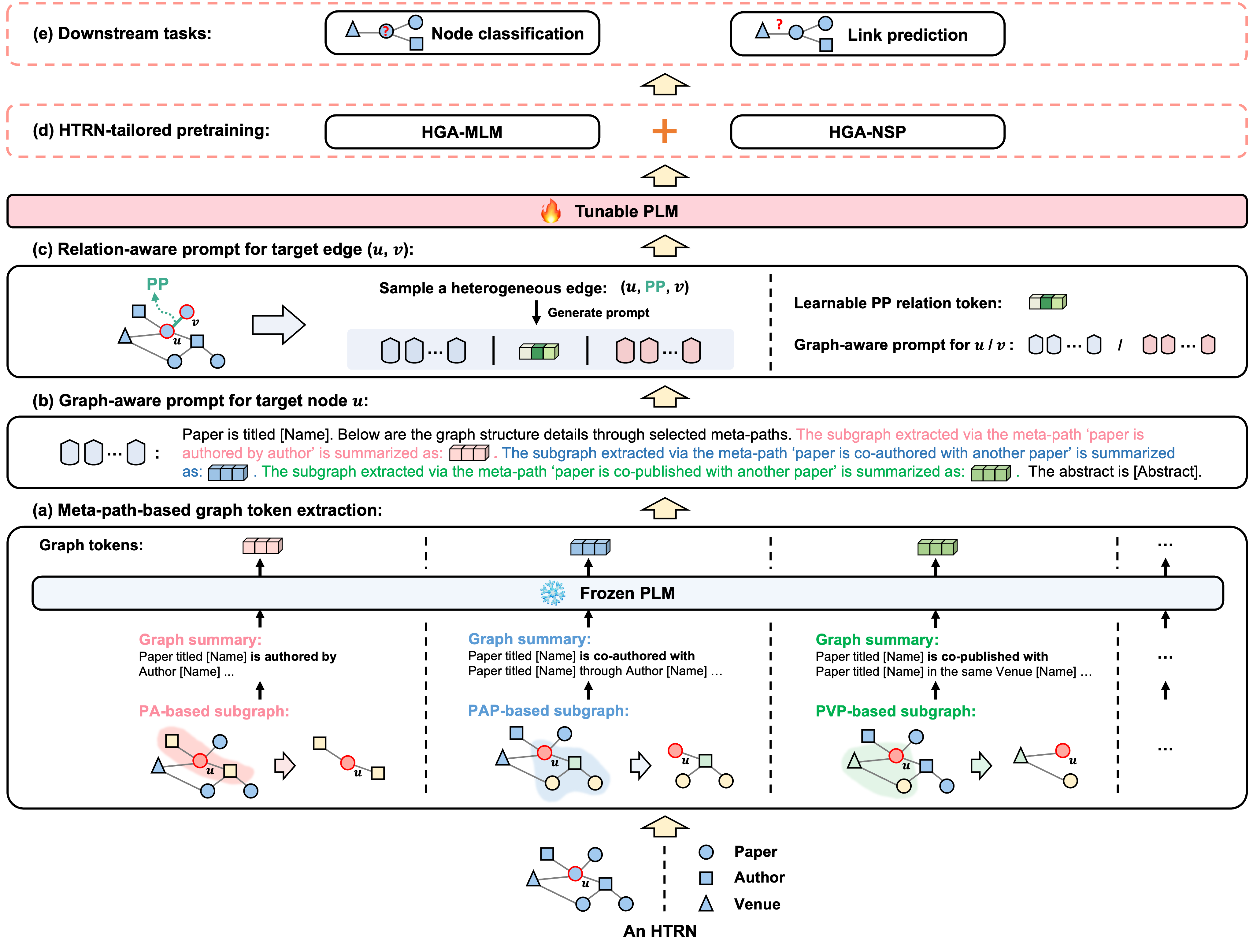}
    \vspace{-3mm}
    \caption{The framework of HierPromptLM. (a) Meta-path-based graph token {\zhu generation}: extract meta-path-based subgraphs from the HTRN, create meaningful textual summaries of these subgraphs, and {\zhu distill} graph tokens using a frozen {\qywww PLM}. (b) 
    Graph-aware prompt: for each target node, {\zhu generate a text sequence by integrating its own text data with meta-path-based graph tokens and their corresponding descriptions. 
    (c) Relation-aware prompt: for each 
    target edge,
    generate a text sequence by combining each node's graph-aware prompt with a learnable relation token that connects them.} (d) HTRN-tailored pretraining: fine-tune a PLM with the relation-aware prompt using two specialized pretraining tasks. (e) Downstream tasks.}
    \vspace{-3mm}
    \label{fig:model}
\end{figure*}

\vspace{-2mm}
\subsection{Graph-aware Prompt}



{\wwwqy 

{\qy To capture each node's inherent heterogeneous structures within HTRNs}, we design a graph-aware prompt that leverages meta-path-based subgraph contexts surrounding the target node for structural augmentation. {\zlwww The key processes involved in our approach are described as follows.}
\subsubsection{Meta-path-based subgraph.}
Inspired by the effectiveness of meta-paths in capturing local context~\cite{sun2011pathsim,wang2019heterogeneous}, we introduce them to model the heterogeneous structural information of each target node in HTRNs. {\zlwww Specifically,} in HTRNs, nodes are {\zlwww connected through} various semantic paths, known as meta-paths~\cite{wang2019heterogeneous}.

\noindent \textbf{Definition 4.1. Meta-path.} 
A meta-path $m$ is a path $c_1 \xrightarrow{r_1} c_2 \xrightarrow{r_2} \dots \xrightarrow{r_k} c_{k+1}$ (abbreviated as $c_1c_2\dots c_{k+1}$), which describes a composite relation $R$ connecting two endpoint nodes through a series of intermediate nodes and relations, where $c_i$ denotes the node type and $r_i$ denotes the relation type.


Built upon the concept of meta-path, we first introduce a novel structure named the meta-path-based subgraph, which is formally defined as follows.

\noindent \textbf{Definition 4.2. Meta-path-based Subgraph}. Given a node $u$ and a meta-path $m$ in HTRNs, the meta-path-based subgraph $G_{m}(u)$ consists of node $u$, its meta-path-based neighbors (which refer to nodes connected to $u$ via meta-path $m$), and 
all intermediate nodes along the path. 

Taking Figure~\ref{fig:model} as an example, for a target node $u$ with a predefined meta-path $PAP$, the extracted $PAP$-based subgraph of $u$ is highlighted with blue shading. {\qywww This subgraph includes the target node (in red), its  meta-path-based neighbors (in yellow), and the intermediate nodes connecting them (in green).} 
This structure captures multi-level information by extracting a subgraph centered on each target node, based on a pre-defined meta-path. It incorporates neighbors at different distances, {\qy  each carrying unique meanings}: the target node at distance 0 reflects its own characteristics, the intermediate nodes at distance 1 indicate authorship associations, and the {\qywww meta-path-based} neighbors at distance 2 signify co-authorship connections. By incorporating neighbors at multiple levels, the model gains a comprehensive view of the target node, from its intrinsic characteristics to broader structural context within HTRNs. 


For each target node, different meta-paths can be applied to divide its local structure into distinct subgraphs, each representing a unique semantic aspect. For example, using predefined meta-paths {\zhu such as} $PA$ and $PVP$, the corresponding subgraphs can be extracted: the $PA$-based subgraph (shaded in {\zhu red}) reflects collaboration, while the $PVP$-based subgraph (shaded in green) highlights shared research topics. Each meta-path-based subgraph provides unique semantic insight, and together, all these subgraphs capture the broader context of the target node within HTRNs, enhancing the target node's representation.



\noindent \textbf{Meta-path-based graph summarization.}  
{\zhu To integrate these meta-path-based subgraphs into a textual space, we employ a subgraph program function $\mathcal{P(\cdot)}$ to textualize the subgraphs into graph summaries. Specifically, for each subgraph, this function generates a text sequence by combining the textual information of nodes in the subgraph with the meta-path's semantics. For example, as shown in Figure~\ref{fig:model}(a), consider a target paper node $u$ with information $\Phi(u) = \{$Meta-path: $PA$; $PA$ semantics: \textit{``is authored by''};   $PA$-based subgraph $G_{PA}(u)$: $\{ u$,  \textit{Author1, Author2}  $\} \}$.  
By applying the function $\mathcal{P(\cdot)}$, we  generate the textual summary for the $PA$-based subgraph of $u$, i.e., $\mathcal{P}(u, G_{PA}(u)) =$ $<$ Paper titled [Name of $u$] is authored by Author [Name of Author1], Author [Name of Author2] $>$. 


}

Note that our method is easily extendable to accommodate more complex meta-paths and efficiently generates meta-path-based graph summaries. This process is highly parallelizable, allowing for the simultaneous generation of a large number of meta-path-based subgraph summaries for each node during the preprocessing stage.

{\wwwqy \subsubsection{Meta-path-based graph tokens.}
{\qy 
{\zhu Directly concatenating {\zhu various} meta-path-based subgraph summaries with the target node's own textual information often results in lengthy inputs, which can be challenging for PLMs to handle due to token length limitations~\cite{bi2024lpnl}.}} {\wwwqyz To mitigate this, we first employ a novel hierarchical structure to   
{\zlwww {\qy distill these meta-path-based subgraph summaries} into concise semantic tokens, referred to as graph tokens in this paper,} before their integration. Specifically, 
{\zlwww we feed the meta-path-based subgraph summaries, which contain rich structural information expressed through abundant natural language, into a frozen PLM to generate concise graph tokens without {\qywww any additional training}, as shown in Figure~\ref{fig:model}(a).}
This process can be efficiently executed offline and, similar to the previous step, supports parallel processing. 
}

\subsubsection{{\qywww Integration of textual and {\zhu heterogeneous} graph structural information}}
{\zhu 
After generating the meta-path-based graph tokens, which act as specialized soft prompts, they are merged with the target node's own textual information via a graph-aware program function $\mathcal{F}(\cdot)$ to construct a graph-aware prompt. Specifically, given a target paper node $u$ with information $\phi(u)=\{ PA$-based graph token: $\mathcal{T}_G(u,PA)$, Token  descriptions: \textit{``The subgraph extracted via the meta-path `paper is authored by author' is summarized as:''} $\}$. By using the function $\mathcal{F}(\cdot)$, we can derive the graph-aware prompt for $u$ as: \textbf{$\mathcal{F}(u) = $} $<$ Paper is named [Name of $u$].  The subgraph extracted via the meta-path `paper is authored by author' is summarized as: [$\mathcal{T}_G(u,PA)$] $>$. 
Additional node-specific information, such as abstract can be appended to the prompt if applicable, e.g., {\zhu ``The abstract is: [Abstract of $u$]''}. Formally, the graph-aware prompt for the target node can be defined as:

\textbf{$\mathcal{F}(u) = $} $<$ [Text Data] [Token Descriptions] [Graph Token] $>$, 

\noindent where ``Text Data'' refers to the target node's textual information, ``Graph Token'' represents the distilled graph tokens, and ``Token Descriptions'' provides the semantic descriptions of meta-path-based graph tokens. 
This structured approach, as shown in Figure~\ref{fig:model}(b), offers a flexible yet consistent framework for various node types, enabling the model to seamlessly integrate rich textual and heterogeneous graph structural information into a unified textual space. 

}

\subsection{Relation-aware Prompt}
{\zhu Except for the node-level modeling using the graph-aware prompt framework, it is also important to further capture the heterogeneous structural information in HTRNs more comprehensively from a complementary edge-level perspective, as edges in HTRNs represent critical relations between various types of nodes, such as authorship, publication, and citation. {\qy These relations provide essential context for understanding node interactions within HTRNs.} 

To address such heterogeneous edge-level information in HTRNs and effectively integrate it with rich textual information into a unified representation space, we propose the relation-aware prompt module.} {\wwwqyz 
This module is designed to capture the heterogeneity of relations within HTRNs by introducing a learnable relation token. This token is strategically placed between the graph-aware prompts of connected nodes to form a comprehensive relation-aware prompt.}
{\qy Specifically, given a node pair $(u,v)$ connected by a relation $r\in \mathcal{R}$, the relation-aware prompt $\mathcal{H}(u,r,v)$ is constructed as:}



{\zhu $\mathcal{H}(u,r,v) = < $ [$\mathcal{F}(u)$] [Relation Token of $r$] [$\mathcal{F}(v)$] $>$, 
}


\noindent where $\mathcal{F}(u)$ and $\mathcal{F}(v)$ denote the graph-aware prompt for node $u$ and $v$, respectively. 
{\wwwqyz Taking  Figure~\ref{fig:model}(c) as an example, {\zlwww assuming} that a pair of nodes $(u,v)$ connected by a $PP$ relation, representing a citation link, {\zlwww is sampled as the target heterogeneous edge}. In this case, $PP$ relation is first encoded as a {\zlwww relation} token and then placed between the graph-aware prompts of nodes $u$ and $v$, constructing a relation-aware prompt that seamlessly integrates the citation interaction context.}


\subsection{HTRN-tailored Pretraining}
Despite the broad generalizability of PLMs, fine-tuning is essential for adapting their capabilities to the unique characteristics of HTRNs. 
{\wwwqyz However, existing methods simply fine-tune PLMs using basic graph reconstruction~\cite{jin2023heterformer} or binary matching {\zlwww tasks}~\cite{zou2023pretraining}, neglecting the heterogeneity and unique semantics of relations, as well as the intricate interactions between text data and graph structures. 
Additionally, these tasks are not specifically designed for PLMs, limiting their ability to fully leverage the strengths of PLMs in handling complex heterogeneous graph structures combined with rich textual information within HTRNs.}
To deal with it, we design two specialized pretraining tasks for HTRNs: HGA-NSP and HGA-MLM. Specifically, HGA-NSP extends the traditional NSP~\cite{kenton2019bert} by capturing different relations with distinct semantics in HTRNs, while HGA-MLM enhances the traditional  MLM~\cite{kenton2019bert} by integrating natural language with graph tokens, enabling a more comprehensive understanding of both textual and heterogeneous structural information, as well as their interactions within HTRNs.

\subsubsection{HGA-NSP}
The HGA-NSP task adapts the traditional NSP for the HTRN context. Specifically, given a pair of nodes $(u,v)$ with a specific relation $r$, this task is to predict whether node $v$ connects node $u$ via the relation $r$. To generate positive samples, we form tuples $(u,r,v)$ where $u$ and $v$ are connected by the relation $r$ within the HTRN. For each positive sample, we generate the corresponding negative samples with a negative sampling ratio $nsr$ (i.e., $nsr$ negative samples per positive sample). Specifically, we create $nsr$ tuples $(u,r,v')$ where $v'$ has the same type as $v$, but not connected to $u$. The model calculates a probability $p$ for each sample, indicating how likely the relation between $u$ and $v$ (or $v'$) is valid. The loss function is expressed as:
\begin{equation}
\mathcal{L}_{HGA\text{-}NSP} = - \frac{1}{N} \sum_{i=1}^{N} [ y_i \log(p_i) + (1 - y_i) \log(1 - p_i) ],
\end{equation}
where $N$ is the total number of samples (both positive and negative), $y_i$ is 1 for positive samples and 0 for negative samples.

\subsubsection{HGA-MLM}
The HGA-MLM task fine-tunes the model by predicting masked tokens {\wwwqyz from} both {\qy text} and graph tokens {\wwwqyz within} the relation-aware prompt. Specifically, for each relation-aware prompt, a specified percentage of tokens are randomly masked according to the masking ratio $mr${\wwwqyz, and the} model needs to predict the masked tokens based on the remaining context. {\qy Different from traditional MLM tasks in NLP, which rely solely on text data, the context in HGA-MLM involves both text data and heterogeneous graph structures. Hence, HGA-MLM allows the model to jointly grasp both textual and heterogeneous structural information, along with their interactions embedded within the relation-aware prompt.} The loss function is defined as:
\begin{equation}
\mathcal{L}_{HGA\text{-}MLM} = - \frac{1}{M} \sum_{i=1}^{M} \log P(x_i | X_{\backslash i}),
\end{equation}
where $M$ is the number of masked tokens, $x_i$ is the original token at position $i$, $X_{\backslash i}$ represents the input sequence with the $i$-th token masked. $P(x_i | X_{\backslash i})$ is the model’s predicted probability for the correct token.

\subsubsection{Overall loss}
The final loss integrates both HGA-MLM and HGA-NSP {\zlwww into a multi-task learning framework}, enabling the model to simultaneously learn  heterogeneous node-level and edge-level information within HTRNs. The overall loss is defined as:
\begin{equation}
\mathcal{L}_{total} = \mathcal{L}_{HGA\text{-}NSP} + \mathcal{L}_{HGA\text{-}MLM}  
\end{equation}

After fine-tuning the PLM using the combined loss {\zlwww function}, we can {\zlwww leverage it to generate representations for a given HTRN}, including both node embeddings from the graph-aware prompt and edge embeddings from the relation-aware prompt, {\zlwww which can be utilized to facilitate various downstream applications, such as node classification and link prediction.}
}}

\section{Experiments}
\label{section:experiment}

\begin{table*}[thb]
\caption{{\zhu Dataset statistics.}}
\vspace{-3mm}
\label{tab:data}
\centering
\resizebox{0.93\textwidth}{!}{\begin{tabular}{c|ccccccc}
\toprule
Dataset & Objects ($\#$) & $\#$Object & Relations & $\#$Relation & $\#$Label Type & $\#$Labeled Object \\ \hline
DBLP & $P$(53,614), $A$(10,279), $V$(3,524) & 67,417 & $P\rightleftharpoons P, P\rightleftharpoons A, P\rightleftharpoons V$ & 149,545 & 4 & 53,614 \\
OAG & $P$(41,632), $A$(20,366), $F$(20), $I$(1,790) & 63,808 & $P\rightleftharpoons P, P\rightleftharpoons A, P\rightleftharpoons F, A\rightleftharpoons I$ & 240,219 & 5 & 41,632 \\
\bottomrule
\end{tabular}
}
\end{table*}

In this section, we conduct extensive experiments to answer the following research questions:
\begin{itemize}[leftmargin=*]
    \item {\zlwww \textbf{RQ1:}} Can HierPromptLM outperform all baselines across various downstream tasks?
    \item {\zlwww \textbf{RQ2:}} How do different modules of HierPromptLM contribute to enhancing the model performance?
    \item {\zlwww \textbf{RQ3:}} How does HierPromptLM perform in a training-free setting to validate its generalization {\qywww capability}?
    \item {\zlwww \textbf{RQ4:}} Can HierPromptLM be extended to different PLM-based backbones, and how does it perform with these extensions?
    \item {\zlwww \textbf{RQ5:}} How do different hyper-parameter settings impact the performance of HierPromptLM ({\zlwww provided in Appendix~\ref{sec:add_para} due to page limitations})?
\end{itemize}

\subsection{Experimental Settings}
\subsubsection{Datasets.} 
We conduct experiments on two publicly available real-world HTRN datasets (i.e. DBLP\footnote{\url{https://originalstatic.aminer.cn/misc/dblp.v12.7z}} and OAG\footnote{\url{https://github.com/UCLA-DM/pyHGT/}}), which are employed in previous works~\cite{jin2023heterformer,zou2023pretraining}. 
The main statistics of datasets are summarized in Table~\ref{tab:data} and the details are provided in Appendix~\ref{sec:dataset}.

\subsubsection{Baselines.} To verify the effectiveness of our model, we compare our HierPromptLM with three groups of baselines: (1) HG-based methods include  shallow model-based methods containing ComplEx~\cite{trouillon2016complex}, HIN2Vec~\cite{fu2017hin2vec} and M2V~\cite{dong2017metapath2vec}, and deep model-based methods involving ie-HGCN~\cite{yang2021interpretable} and SHGP~\cite{yang2022self}; (2) PLM methods including Bert~\cite{kenton2019bert} and Albert~\cite{chi2021audio}; (3) HTRN-based methods containing Heterformer~\cite{jin2023heterformer} and THLM~\cite{zou2023pretraining}. 
{\zhu Specifically, HG-based methods
are designed for heterogeneous graphs without considering text data. Following~\cite{zou2023pretraining}, we extend these models by using BERT to first extract the textual information for each node as its initial embedding, resulting in the models referred to as ComplEx+Bert, HIN2Vec+Bert, M2V+Bert, ie-HGCN+Bert, SHGP+Bert. In contrast, PLM methods focus solely on text data, whereas HTRN-based methods are specifically designed for HTRNs, considering both textual and structural information.} 
More details about reproducibility can be found in Appendix~\ref{sec:rep}.

\begin{table}[h]
\vspace{-3mm}
  \caption{Overall evaluation {\zhu on} node classification. Tabular results are in percent; the best results are highlighted in bold; the \underline{underlined} results indicate the second-best performance.}
  \vspace{-2.5mm}
  \label{tab:ncla}
  \setlength\tabcolsep{4pt}
  \centering
  \resizebox{0.424\textwidth}{!}{\begin{tabular}{ccccc}
  \toprule
  \multirow{2}{*}{Methods}  & \multicolumn{2}{c}{DBLP} & \multicolumn{2}{c}{OAG}  \\
  & Micro-F1 & Macro-F1 & Micro-F1 & Macro-F1 \\ \hline
  {\zhu ComplEx+Bert} & 55.55 & 26.40 &  64.33 & 57.59 \\
   & ($\pm$0.13) & ($\pm$0.38) &  ($\pm$0.02) & ($\pm$0.03) \\
  {\zhu HIN2Vec+Bert} & 54.62 & 17.66 &  28.87 & 23.71 \\
  & ($\pm$0.00) & ($\pm$0.00) &  ($\pm$0.00) & ($\pm$0.00) \\
  {\zhu M2V+Bert} & 68.32 & 56.26 & 86.16 & 84.01 \\
  & ($\pm$0.32) & ($\pm$0.96) & ($\pm$0.09) & ($\pm$0.09) \\ \hline
  {\zhu ie-HGCN+Bert} & 53.36 & 18.29 &  29.74 & 27.08 \\
  & ($\pm$2.51) & ($\pm$1.33) &  ($\pm$6.02) & ($\pm$5.59) \\
  {\zhu SHGP+Bert} & 52.30 & 20.89 & 34.85  & 24.40 \\ 
  & ($\pm$2.58) & ($\pm$3.05) & ($\pm$11.06)  & ($\pm$10.86) \\ \hline
  Bert & 81.67 & 77.30 & 85.99  & 85.78 \\
  & ($\pm$1.53) & ($\pm$2.29) & ($\pm$1.52)  & ($\pm$1.56) \\
  Albert & 79.08 & 73.82 & 85.01 & 84.86 \\ 
  & ($\pm$1.57) & ($\pm$2.87) &  ($\pm$1.07) & ($\pm$1.17) \\ \hline
  Heterformer & \underline{86.40} & \underline{83.10} & 93.82 & 90.61 \\
  & ($\pm$0.08) & ($\pm$0.12) & ($\pm$0.28) & ($\pm$0.29) \\
  THLM & 85.21 & 82.12 & \underline{94.83}  & \underline{92.18} \\
  & ($\pm$1.07) & ($\pm$1.46) & ($\pm$0.56) & ($\pm$0.87) \\ \hline
  {\zhu HierPromptLM} & \textbf{90.36} & \textbf{88.15} & \textbf{96.24}  & \textbf{94.43} \\
  & ($\pm$0.07) & ($\pm$0.09) & ($\pm$0.06)  & ($\pm$0.09) \\ \hline
  Improvement & \textbf{4.58\%} & \textbf{6.08\%} & \textbf{1.49\%} & \textbf{2.44\%}   \\ 
  \bottomrule
  \end{tabular}}
  \vspace{-3.5mm}
\end{table}

\subsection{Node Classification (RQ1)}
\noindent \textbf{Settings.} Node classification aims to assign categories to nodes within a network. Following \cite{hu2020heterogeneous}, we train a separate linear Support Vector Machine (LinearSVM) \cite{fan2008liblinear} {\wwwqyz using the node embeddings generated by our model as input, with the corresponding node labels serving as the classification targets.}
The ratio of the nodes used for training, validation, and testing is 70\%, 10\%, and 20\% in all datasets. 
All models are trained for 10 times and, the mean and standard variance of test performance are reported. Micro-F1 and Macro-F1 scores are employed to evaluate the effectiveness following ~\cite{zou2023pretraining, jin2023heterformer}.

\noindent \textbf{Results.} 
The overall experimental results are shown in Table~\ref{tab:ncla}. As observed, HierPromptLM consistently surpasses all baselines in node classification across both datasets, demonstrating its superior effectiveness. Specifically, HierPromptLM exhibits an improvement of up to 4.58\% in Micro-F1 and 6.08\% in Macro-F1 over the top-performing baselines on the DBLP dataset. The main reasons for the observed improvement may include: (1) HierPromptLM jointly models both textual and heterogeneous structural information within a unified representation space, {\qy enabling to fully capture their {\zhu critical} interactions within HTRNs}. (2) Two HTRN-tailored pretraining tasks facilitate a deeper integration and understanding of both the textual and heterogeneous structural information, thereby enhancing the overall model performance. Additionally, HTRN-based methods surpass PLM methods, highlighting the importance of integrating both textual and structural information to achieve superior performance. {\zhu However, PLM methods still significantly outperform HG-based models combined with BERT, as the naive integration of text data and structures (i.e., simply using BERT embeddings as initial features) is often ineffective, leading to worse performance compared to text-only models for node classification.
}





\subsection{Link Prediction (RQ1)} \noindent \textbf{Settings.} Link prediction aims to predict missing edges in a network. 
{\wwwqyz In HTRNs, there are various heterogeneous edges, and missing links can belong to any of these types. Therefore, to comprehensively evaluate the model performance, it is essential to consider multiple relation types simultaneously (e.g., paper-paper, paper-author, paper-venue {\qy for the DBLP dataset}, and paper-paper, paper-author, paper-field, author-institution {\qy for the OAG dataset}), which enables a full assessment of the model's ability to capture {\qy diverse  interactions inherent in HTRNs.}}
Furthermore, we also conduct link prediction for each relation type on the DBLP dataset to further assess its effectiveness in capturing specific relations within HTRNs, and the detailed results are provided in Appendix~\ref{sec:add_link}.  Following~\cite{zou2023pretraining}, we apply a standard sampling strategy for link prediction, using 10\% of edges for training and 40\% for testing across both datasets. Evaluation metrics, including ROC-AUC, PR-AUC, and F1 scores, are used following previous work~\cite{liu2022aspect}.

\begin{table}[h]
\vspace{-2mm}
  \caption{Overall evaluation {\zhu on} link prediction. 
  Tabular results are in percent; the best results are highlighted in bold; the \underline{underlined} results indicate the second-best performance.
  }
  \label{tab:link}
  \vspace{-2.5mm}
  \setlength\tabcolsep{1.5pt}
  \centering
  \resizebox{0.478\textwidth}{!}{\begin{tabular}{ccccccc}
  \toprule
  \multirow{2}{*}{Methods} & \multicolumn{3}{c}{DBLP} & \multicolumn{3}{c}{OAG} \\ 
  & ROC-AUC & PR-AUC & F1 & ROC-AUC & PR-AUC & F1 \\ \hline
  {\zhu ComplEx+Bert} & 66.38 & 61.46 & 63.03 & 77.86 & 72.77 & 76.44 \\
  & ($\pm$0.03) & ($\pm$0.06) & ($\pm$0.16) & ($\pm$0.02) & ($\pm$0.07) & ($\pm$0.04) \\
  {\zhu HIN2Vec+Bert} & 50.55 & 50.28 & 50.37 & 66.96 & 64.44 & 55.44 \\
   & ($\pm$0.00) & ($\pm$0.00) & ($\pm$0.00) & ($\pm$0.00) & ($\pm$0.00) & ($\pm$0.00) \\
  {\zhu M2V+Bert} & \underline{78.36} & \underline{72.57} & \underline{77.93} & \underline{84.04} & \underline{80.11} &  \underline{83.09}\\ 
  & ($\pm$0.16) & ($\pm$0.48) & ($\pm$0.56) & ($\pm$0.27) & ($\pm$0.87) & ($\pm$0.18) \\ \hline
  {\zhu ie-HGCN+Bert} & 71.09 & 66.51 & 68.16 & 68.27 & 63.00 & 67.45 \\
  & ($\pm$3.83) & ($\pm$4.13) & ($\pm$7.68) & ($\pm$3.88) & ($\pm$3.87) & ($\pm$3.77) \\ 
  {\zhu SHGP+Bert} & 68.02 & 63.74 & 66.55 & 80.05 & 73.84 & 81.48 \\ 
  & ($\pm$6.51) & ($\pm$6.63) & ($\pm$8.15) & ($\pm$3.98) & ($\pm$5.87) & ($\pm$2.38) \\  \hline
  Bert & 71.12 & 67.77 & 72.39 & 80.42 & 76.59 & 80.84 \\
   & ($\pm$10.51) & ($\pm$9.26) & ($\pm$3.01) & ($\pm$5.69) &  ($\pm$5.69) & ($\pm$8.33) \\ 
  Albert & 70.34 & 65.94 & 71.96 & 83.39 & 79.22 & 82.88 \\ 
  & ($\pm$8.33) & ($\pm$7.61) & ($\pm$2.96) & ($\pm$2.37) & ($\pm$4.02) & ($\pm$0.94) \\  \hline
  Heterformer & 72.04 & 70.27 & 75.46 & 79.45 & 76.49 & 80.91 \\
  & ($\pm$0.91) & ($\pm$1.09) & ($\pm$0.77) & ($\pm$0.25) & ($\pm$0.52) & ($\pm$0.35) \\ 
  THLM & 72.60 & 68.18 & 70.13 & 79.53 & 73.35 & 79.76 \\ 
  & ($\pm$3.36) & ($\pm$4.18) & ($\pm$4.06) & ($\pm$1.28) & ($\pm$1.64) & ($\pm$1.26) \\  \hline
  {\zhu HierPromptLM} & \textbf{85.56} & \textbf{80.44} & \textbf{85.56} & \textbf{89.24} & \textbf{84.92} & \textbf{89.27} \\ 
  & ($\pm$0.04) & ($\pm$0.01) & ($\pm$0.05) & ($\pm$0.01) & ($\pm$0.07) & ($\pm$0.01) \\  \hline
  Improvement & \textbf{9.19\%} & \textbf{10.84\%} & \textbf{9.79\%} & \textbf{6.19\%} & \textbf{6.00\%} & \textbf{7.44\%} \\
  \bottomrule
  \end{tabular}}
  \vspace{-3.5mm}
\end{table}

\noindent \textbf{Results.}
As shown in Table~\ref{tab:link}, our HierPromptLM shows superior performance in link prediction across both datasets, {\wwwqyz achieving a significant 9.19\% gain in ROC-AUC, 10.84\% improvement in PR-AUC, and 9.79\% enhancement in F1 }over the second-best performance on the DBLP dataset. 
{\wwwqyz This can be attributed to two key factors: (1) a learnable relation token that explicitly models the heterogeneity of relations within HTRNs, and (2) the relation-aware prompt, which employs a hierarchical architecture to {\qy effectively models textual information, heterogeneous node-level and edge-level graph structures, as well as their inherent interactions all within a unified representation space}.}
{\zhu However, PLM methods still significantly outperform HG-based models combined with BERT, consistent with our observations in node classification task. 
} 
{\zhu M2V-Bert} stands out among all baselines because it can effectively employ meta-paths to capture heterogeneous structures and trace the meta-paths to understand heterogeneous relations between nodes in HTRNs. Additionally, HTRN-based methods show better performance than PLM methods on the DBLP dataset but underperform on the OAG dataset. This discrepancy arises because the DBLP dataset has a simpler graph structure, while the OAG dataset is more complex, making it harder to model both structural and textual information simultaneously. 
{\zhu Specifically,} HTRN-based methods process textual and structural information separately, failing to capture their {\zhu critical} interactions and {\zhu often necessitating an} extra alignment that may introduce misleading signals.
In contrast, HierPromptLM effectively integrates both types of information within a unified {\zhu textual} space, achieving the best performance across both datasets.

\begin{figure*}[thb]
    \centering     \includegraphics[width=0.95\textwidth]{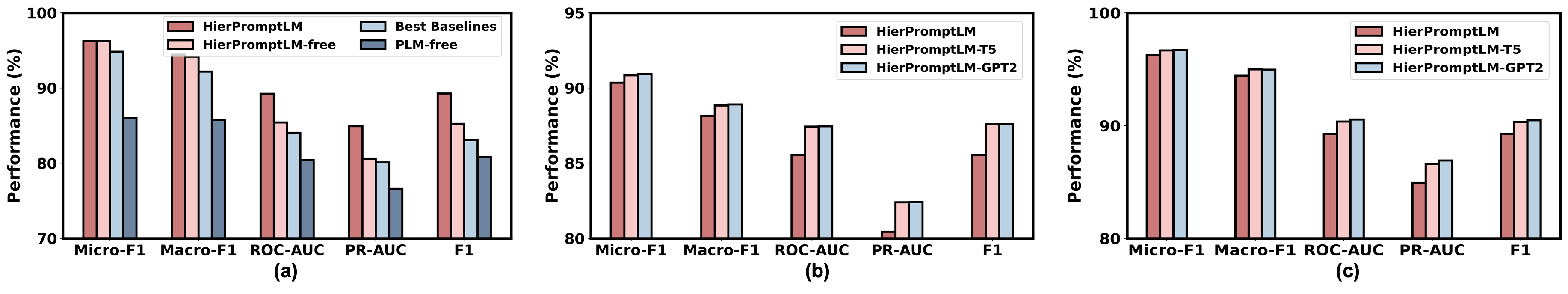}
    \vspace{-4.5mm}
    \caption{(a) Training-free extension on OAG. (b) PLM backbone extension on DBLP. (c) PLM backbone extension on OAG.}
    \label{fig:ex1}
    \vspace{-5mm}
\end{figure*}

\subsection{Ablation Study (RQ2)}
\label{sec:ablation}
To assess the impact of different components in our proposed pure PLM-based framework on overall performance, we conduct ablation studies by removing key modules from HierPromptLM across both datasets. Specifically, we focus on four main modules: (1) \textbf{w/o HGA-MLM} removes the HGA-MLM task; (2) \textbf{w/o HGA-NSP} {\qywww eliminates} the HGA-NSP task; (3) \textbf{w/o GraphToken} {\qywww deletes} the meta-path-based graph tokens from the graph-aware prompt for each target node, relying solely on text data in the graph-aware prompt; (4) \textbf{w/o RelationToken} {\qywww excludes} the relation token from the relation-aware prompt. The results for node classification and link prediction on the OAG dataset are presented in Table~\ref{tab:oag_ablation}, with additional details for the DBLP dataset available in Appendix~\ref{sec:dblp_ablation}. 

Based on the results, we have the following observations: (1) Removing the HGA-MLM task results in a significant performance drop in both {\wwwqyz node classification and link prediction}, highlighting its critical role in capturing both textual and heterogeneous structural information within HTRNs, which is essential for effective representation learning. (2) HierPromptLM consistently outperforms w/o HGA-NSP, confirming the benefits of this tailored pretraining task, which considers the heterogeneity of relations in HTRNs. These findings demonstrate the necessity of designing specific pretraining tasks for fine-tuning PLMs on HTRNs. (3) HierPromptLM shows superior performance compared {\zlwww with} w/o GraphToken across both {\qywww node classification and link prediction}, demonstrating the strength of meta-path-based subgraphs in enhancing node representations by augmenting structural information in HTRNs. (4) The removal of the relation token leads to the decrease in performance, underscoring the importance of modeling heterogeneous relations with a learnable embedding that captures the edge-level heterogeneity within HTRNs {\zlwww explicitly}, thereby improving representation learning on HTRNs.

\begin{table}[thb]
\vspace{-2mm}
  \caption{Ablation study on OAG. Tabular results are in percent; the best results are highlighted in bold.}
  \label{tab:oag_ablation}
  \vspace{-3mm}
  \setlength\tabcolsep{2pt}
  \centering
  \resizebox{0.46\textwidth}{!}{\begin{tabular}{cccccc}
  \toprule
    \multirow{2}{*}{Methods} & \multicolumn{2}{c}{Node classification} & \multicolumn{3}{c}{Link Prediction} \\ 
   & Micro-F1 & Macro-F1 & ROC-AUC & PR-AUC & F1 \\ \hline

  w/o HGA-MLM & 83.07 & 78.93 & 76.85 & 71.35 & 76.16 \\ 
  & ($\pm$1.42) & ($\pm$1.89) & ($\pm$0.90) & ($\pm$2.13) & ($\pm$2.71) \\
  w/o HGA-NSP & 93.97 & 91.13 & 84.47 & 80.26 & 83.84 \\ 
  & ($\pm$0.09) & ($\pm$0.10) & ($\pm$0.38) & ($\pm$1.74) & ($\pm$0.78) \\
  w/o GraphToken & 93.35 & 90.63 & 77.56 & 71.56 & 77.32 \\ 
  & ($\pm$0.10) & ($\pm$0.12) & ($\pm$0.34) & ($\pm$0.70) & ($\pm$0.27) \\
  w/o RelationToken & 96.16 & 94.33 & 88.93 & 84.72 & 88.89 \\ 
  & ($\pm$0.07) & ($\pm$0.12) & ($\pm$0.01) & ($\pm$0.05) & ($\pm$0.01) \\ \hline

  {\zhu HierPromptLM} & \textbf{96.24} & \textbf{94.43} & \textbf{89.24} & \textbf{84.92} & \textbf{89.27} \\ 
  & ($\pm$0.06) & ($\pm$0.09) & ($\pm$0.01) & ($\pm$0.07) & ($\pm$0.01) \\  
  \bottomrule
  \end{tabular}
  }
  \vspace{-3mm}
\end{table}

\vspace{-2mm}
{\qywww \subsection{Evaluation of Training-free Setting (RQ3)} 
\label{sec:oag_free}
To evaluate the generalization capability of HierPromptLM, we introduce a training-free variant called \textbf{HierPromptLM-free}, where we skip fine-tuning on a tunable PLM {\wwwqyz using HGA-MLM and HGA-NSP} after generating the relation-aware prompt, and instead directly feed {\zhu this} prompt into a frozen PLM {\wwwqyz without any additional training}. We apply this approach to node classification and link prediction tasks across both datasets{\wwwqyz, following the same settings as discussed above.} For comparison, we include the top-performing results among all baselines (referred to as \textbf{Best Baselines}) and an existing {\zhu pretrained} language model (Bert) without training (referred to as \textbf{PLM-free}). The results for the OAG dataset are shown in  Figure~\ref{fig:ex1}(a), with additional results for the DBLP dataset provided in Appendix~\ref{sec:dblp_free}. As observed, though HierPromptLM-free performs slightly below HierPromptLM, it still consistently and significantly outperforms both Best Baselines and PLM-free. This demonstrates that HierPromptLM has strong generalization capability, even in a training-free setting. {\zlwww HierPromptLM-free offers a significant advantage by reducing the computational cost and time required for fine-tuning, while still retaining most of the performance benefits. This makes it especially suitable for scenarios where fine-tuning large models is impractical, {\wwwqyz such as when model parameters are inaccessible or fine-tuning demands significant computational resources, }yet high-quality results remain essential.}}



\vspace{-2mm}
\subsection{Extension of PLM-based Backbones (RQ4)} 
To further explore our proposed HierPromptLM, we replace the frozen PLM backbone (Bert-base{\zhu \footnote{\url{https://huggingface.co/google-bert}}}) with other PLMs, including T5-base\footnote{\url{https://huggingface.co/google-t5/t5-base}} (220M parameters, \cite{2020t5}) and GPT2-small\footnote{\url{https://huggingface.co/openai-community/gpt2}} (124M parameters, \cite{radford2019language}), for encoding the meta-path-based graph tokens. We opt not to replace the tunable PLM during fine-tuning because fine-tuning large models is time-intensive and can be considered for future work. Instead, frozen PLMs offer the advantage of offline processing, enabling us to work with PLMs of any size. This experiment is intended to assess the flexibility of HierPromptLM by evaluating the impact of swapping in different PLMs {\zhu on its performance}. The results on the DBLP and OAG datasets are presented in Figure~\ref{fig:ex1}(b) and (c), respectively. As observed, both HierPromptLM-T5 and HierPromptLM-GPT2 significantly outperform HierPromptLM, which can be attributed to {\zlwww the fact that} T5 and GPT2 {\zlwww have} more parameters and stronger learning capability than Bert. These results not only validate the enhanced performance with more powerful PLMs but also highlight the flexibility of our {\zlwww proposed framework}. It demonstrates that HierPromptLM can effectively adapt to different PLMs {\wwwqyz and maintain} strong performance.


{\wwwqyz 
\noindent \textbf{Remark: Parameter Analysis (RQ5).} We also investigate the sensitivity of HierPromptLM to three key hyper-parameters: pre-defined meta-paths, the negative sampling ratio and the masking ratio. Due to page limitations, the results for node classification and link prediction across both datasets are provided in Appendix~\ref{sec:add_para}.}

\section{Conclusion}
\label{sec:conclusion}
In this paper, we introduce HierPromptLM, a novel PLM-based method for representation learning on HTRNs, which jointly models textual and structural information, along with their interactions, within a unified representation space. HierPromptLM has two key components: {\zhu an automated textualization mechanism that employs prompt learning to integrate both text data and heterogeneous structures at both node and edge levels into meaningful text sequences,} 
and two novel HTRN-tailored pretraining tasks for fine-tuning. Extensive experiments on two real-world HTRN datasets demonstrate that HierPromptLM outperforms state-of-the-art methods.

\newpage  


\bibliographystyle{ACM-Reference-Format}
\bibliography{ref}


\begin{thebibliography}{50}


\ifx \showCODEN    \undefined \def \showCODEN     #1{\unskip}     \fi
\ifx \showDOI      \undefined \def \showDOI       #1{#1}\fi
\ifx \showISBNx    \undefined \def \showISBNx     #1{\unskip}     \fi
\ifx \showISBNxiii \undefined \def \showISBNxiii  #1{\unskip}     \fi
\ifx \showISSN     \undefined \def \showISSN      #1{\unskip}     \fi
\ifx \showLCCN     \undefined \def \showLCCN      #1{\unskip}     \fi
\ifx \shownote     \undefined \def \shownote      #1{#1}          \fi
\ifx \showarticletitle \undefined \def \showarticletitle #1{#1}   \fi
\ifx \showURL      \undefined \def \showURL       {\relax}        \fi
\providecommand\bibfield[2]{#2}
\providecommand\bibinfo[2]{#2}
\providecommand\natexlab[1]{#1}
\providecommand\showeprint[2][]{arXiv:#2}

\bibitem[Bi et~al\mbox{.}(2024)]%
        {bi2024lpnl}
\bibfield{author}{\bibinfo{person}{Baolong Bi}, \bibinfo{person}{Shenghua Liu}, \bibinfo{person}{Yiwei Wang}, \bibinfo{person}{Lingrui Mei}, {and} \bibinfo{person}{Xueqi Cheng}.} \bibinfo{year}{2024}\natexlab{}.
\newblock \showarticletitle{Lpnl: Scalable link prediction with large language models}. In \bibinfo{booktitle}{\emph{Findings of the Association for Computational Linguistics ACL 2024}}. \bibinfo{pages}{3615--3625}.
\newblock


\bibitem[Brown(2020)]%
        {brown2020language}
\bibfield{author}{\bibinfo{person}{Tom~B Brown}.} \bibinfo{year}{2020}\natexlab{}.
\newblock \showarticletitle{Language models are few-shot learners}.
\newblock \bibinfo{journal}{\emph{arXiv preprint arXiv:2005.14165}} (\bibinfo{year}{2020}).
\newblock


\bibitem[Chen et~al\mbox{.}(2024)]%
        {chen2024llaga}
\bibfield{author}{\bibinfo{person}{Runjin Chen}, \bibinfo{person}{Tong Zhao}, \bibinfo{person}{Ajay Jaiswal}, \bibinfo{person}{Neil Shah}, {and} \bibinfo{person}{Zhangyang Wang}.} \bibinfo{year}{2024}\natexlab{}.
\newblock \showarticletitle{Llaga: Large language and graph assistant}.
\newblock \bibinfo{journal}{\emph{arXiv preprint arXiv:2402.08170}} (\bibinfo{year}{2024}).
\newblock


\bibitem[Chi et~al\mbox{.}(2021)]%
        {chi2021audio}
\bibfield{author}{\bibinfo{person}{Po-Han Chi}, \bibinfo{person}{Pei-Hung Chung}, \bibinfo{person}{Tsung-Han Wu}, \bibinfo{person}{Chun-Cheng Hsieh}, \bibinfo{person}{Yen-Hao Chen}, \bibinfo{person}{Shang-Wen Li}, {and} \bibinfo{person}{Hung-yi Lee}.} \bibinfo{year}{2021}\natexlab{}.
\newblock \showarticletitle{Audio albert: A lite bert for self-supervised learning of audio representation}. In \bibinfo{booktitle}{\emph{2021 IEEE Spoken Language Technology Workshop (SLT)}}. IEEE, \bibinfo{pages}{344--350}.
\newblock


\bibitem[Chien et~al\mbox{.}(2021)]%
        {chien2021node}
\bibfield{author}{\bibinfo{person}{Eli Chien}, \bibinfo{person}{Wei-Cheng Chang}, \bibinfo{person}{Cho-Jui Hsieh}, \bibinfo{person}{Hsiang-Fu Yu}, \bibinfo{person}{Jiong Zhang}, \bibinfo{person}{Olgica Milenkovic}, {and} \bibinfo{person}{Inderjit~S Dhillon}.} \bibinfo{year}{2021}\natexlab{}.
\newblock \showarticletitle{Node feature extraction by self-supervised multi-scale neighborhood prediction}.
\newblock \bibinfo{journal}{\emph{arXiv preprint arXiv:2111.00064}} (\bibinfo{year}{2021}).
\newblock


\bibitem[Dong et~al\mbox{.}(2020)]%
        {dong2020autoknow}
\bibfield{author}{\bibinfo{person}{Xin~Luna Dong}, \bibinfo{person}{Xiang He}, \bibinfo{person}{Andrey Kan}, \bibinfo{person}{Xian Li}, \bibinfo{person}{Yan Liang}, \bibinfo{person}{Jun Ma}, \bibinfo{person}{Yifan~Ethan Xu}, \bibinfo{person}{Chenwei Zhang}, \bibinfo{person}{Tong Zhao}, \bibinfo{person}{Gabriel Blanco~Saldana}, {et~al\mbox{.}}} \bibinfo{year}{2020}\natexlab{}.
\newblock \showarticletitle{Autoknow: Self-driving knowledge collection for products of thousands of types}. In \bibinfo{booktitle}{\emph{Proceedings of the 26th ACM SIGKDD International Conference on Knowledge Discovery \& Data Mining}}. \bibinfo{pages}{2724--2734}.
\newblock


\bibitem[Dong et~al\mbox{.}(2017)]%
        {dong2017metapath2vec}
\bibfield{author}{\bibinfo{person}{Yuxiao Dong}, \bibinfo{person}{Nitesh~V Chawla}, {and} \bibinfo{person}{Ananthram Swami}.} \bibinfo{year}{2017}\natexlab{}.
\newblock \showarticletitle{metapath2vec: Scalable representation learning for heterogeneous networks}. In \bibinfo{booktitle}{\emph{Proceedings of the 23rd ACM SIGKDD international conference on knowledge discovery and data mining}}. \bibinfo{pages}{135--144}.
\newblock


\bibitem[El-Kishky et~al\mbox{.}(2022)]%
        {el2022twhin}
\bibfield{author}{\bibinfo{person}{Ahmed El-Kishky}, \bibinfo{person}{Thomas Markovich}, \bibinfo{person}{Serim Park}, \bibinfo{person}{Chetan Verma}, \bibinfo{person}{Baekjin Kim}, \bibinfo{person}{Ramy Eskander}, \bibinfo{person}{Yury Malkov}, \bibinfo{person}{Frank Portman}, \bibinfo{person}{Sof{\'\i}a Samaniego}, \bibinfo{person}{Ying Xiao}, {et~al\mbox{.}}} \bibinfo{year}{2022}\natexlab{}.
\newblock \showarticletitle{Twhin: Embedding the twitter heterogeneous information network for personalized recommendation}. In \bibinfo{booktitle}{\emph{Proceedings of the 28th ACM SIGKDD conference on knowledge discovery and data mining}}. \bibinfo{pages}{2842--2850}.
\newblock


\bibitem[Fan et~al\mbox{.}(2008)]%
        {fan2008liblinear}
\bibfield{author}{\bibinfo{person}{Rong-En Fan}, \bibinfo{person}{Kai-Wei Chang}, \bibinfo{person}{Cho-Jui Hsieh}, \bibinfo{person}{Xiang-Rui Wang}, {and} \bibinfo{person}{Chih-Jen Lin}.} \bibinfo{year}{2008}\natexlab{}.
\newblock \showarticletitle{LIBLINEAR: A library for large linear classification}.
\newblock \bibinfo{journal}{\emph{the Journal of machine Learning research}}  \bibinfo{volume}{9} (\bibinfo{year}{2008}), \bibinfo{pages}{1871--1874}.
\newblock


\bibitem[Fatemi et~al\mbox{.}(2023)]%
        {fatemi2023talk}
\bibfield{author}{\bibinfo{person}{Bahare Fatemi}, \bibinfo{person}{Jonathan Halcrow}, {and} \bibinfo{person}{Bryan Perozzi}.} \bibinfo{year}{2023}\natexlab{}.
\newblock \showarticletitle{Talk like a graph: Encoding graphs for large language models}.
\newblock \bibinfo{journal}{\emph{arXiv preprint arXiv:2310.04560}} (\bibinfo{year}{2023}).
\newblock


\bibitem[Fu et~al\mbox{.}(2017)]%
        {fu2017hin2vec}
\bibfield{author}{\bibinfo{person}{Tao-yang Fu}, \bibinfo{person}{Wang-Chien Lee}, {and} \bibinfo{person}{Zhen Lei}.} \bibinfo{year}{2017}\natexlab{}.
\newblock \showarticletitle{Hin2vec: Explore meta-paths in heterogeneous information networks for representation learning}. In \bibinfo{booktitle}{\emph{Proceedings of the 2017 ACM on Conference on Information and Knowledge Management}}. \bibinfo{pages}{1797--1806}.
\newblock


\bibitem[Fu et~al\mbox{.}(2020)]%
        {fu2020magnn}
\bibfield{author}{\bibinfo{person}{Xinyu Fu}, \bibinfo{person}{Jiani Zhang}, \bibinfo{person}{Ziqiao Meng}, {and} \bibinfo{person}{Irwin King}.} \bibinfo{year}{2020}\natexlab{}.
\newblock \showarticletitle{Magnn: Metapath aggregated graph neural network for heterogeneous graph embedding}. In \bibinfo{booktitle}{\emph{Proceedings of the web conference 2020}}. \bibinfo{pages}{2331--2341}.
\newblock


\bibitem[Guo et~al\mbox{.}(2023)]%
        {guo2023gpt4graph}
\bibfield{author}{\bibinfo{person}{Jiayan Guo}, \bibinfo{person}{Lun Du}, \bibinfo{person}{Hengyu Liu}, \bibinfo{person}{Mengyu Zhou}, \bibinfo{person}{Xinyi He}, {and} \bibinfo{person}{Shi Han}.} \bibinfo{year}{2023}\natexlab{}.
\newblock \showarticletitle{Gpt4graph: Can large language models understand graph structured data? an empirical evaluation and benchmarking}.
\newblock \bibinfo{journal}{\emph{arXiv preprint arXiv:2305.15066}} (\bibinfo{year}{2023}).
\newblock


\bibitem[Hu et~al\mbox{.}(2020)]%
        {hu2020heterogeneous}
\bibfield{author}{\bibinfo{person}{Ziniu Hu}, \bibinfo{person}{Yuxiao Dong}, \bibinfo{person}{Kuansan Wang}, {and} \bibinfo{person}{Yizhou Sun}.} \bibinfo{year}{2020}\natexlab{}.
\newblock \showarticletitle{Heterogeneous graph transformer}. In \bibinfo{booktitle}{\emph{Proceedings of the web conference 2020}}. \bibinfo{pages}{2704--2710}.
\newblock


\bibitem[Jin et~al\mbox{.}(2023a)]%
        {jin2023patton}
\bibfield{author}{\bibinfo{person}{Bowen Jin}, \bibinfo{person}{Wentao Zhang}, \bibinfo{person}{Yu Zhang}, \bibinfo{person}{Yu Meng}, \bibinfo{person}{Xinyang Zhang}, \bibinfo{person}{Qi Zhu}, {and} \bibinfo{person}{Jiawei Han}.} \bibinfo{year}{2023}\natexlab{a}.
\newblock \showarticletitle{Patton: Language model pretraining on text-rich networks}.
\newblock \bibinfo{journal}{\emph{arXiv preprint arXiv:2305.12268}} (\bibinfo{year}{2023}).
\newblock


\bibitem[Jin et~al\mbox{.}(2023b)]%
        {jin2023heterformer}
\bibfield{author}{\bibinfo{person}{Bowen Jin}, \bibinfo{person}{Yu Zhang}, \bibinfo{person}{Qi Zhu}, {and} \bibinfo{person}{Jiawei Han}.} \bibinfo{year}{2023}\natexlab{b}.
\newblock \showarticletitle{Heterformer: Transformer-based deep node representation learning on heterogeneous text-rich networks}. In \bibinfo{booktitle}{\emph{Proceedings of the 29th ACM SIGKDD Conference on Knowledge Discovery and Data Mining}}. \bibinfo{pages}{1020--1031}.
\newblock


\bibitem[Jin et~al\mbox{.}(2021)]%
        {jin2021bite}
\bibfield{author}{\bibinfo{person}{Di Jin}, \bibinfo{person}{Xiangchen Song}, \bibinfo{person}{Zhizhi Yu}, \bibinfo{person}{Ziyang Liu}, \bibinfo{person}{Heling Zhang}, \bibinfo{person}{Zhaomeng Cheng}, {and} \bibinfo{person}{Jiawei Han}.} \bibinfo{year}{2021}\natexlab{}.
\newblock \showarticletitle{Bite-gcn: A new GCN architecture via bidirectional convolution of topology and features on text-rich networks}. In \bibinfo{booktitle}{\emph{Proceedings of the 14th ACM International Conference on Web Search and Data Mining}}. \bibinfo{pages}{157--165}.
\newblock


\bibitem[Kenton and Toutanova(2019)]%
        {kenton2019bert}
\bibfield{author}{\bibinfo{person}{Jacob Devlin Ming-Wei~Chang Kenton} {and} \bibinfo{person}{Lee~Kristina Toutanova}.} \bibinfo{year}{2019}\natexlab{}.
\newblock \showarticletitle{Bert: Pre-training of deep bidirectional transformers for language understanding}. In \bibinfo{booktitle}{\emph{Proceedings of naacL-HLT}}, Vol.~\bibinfo{volume}{1}. Minneapolis, Minnesota, \bibinfo{pages}{2}.
\newblock


\bibitem[Levine et~al\mbox{.}(2021)]%
        {levine2021inductive}
\bibfield{author}{\bibinfo{person}{Yoav Levine}, \bibinfo{person}{Noam Wies}, \bibinfo{person}{Daniel Jannai}, \bibinfo{person}{Dan Navon}, \bibinfo{person}{Yedid Hoshen}, {and} \bibinfo{person}{Amnon Shashua}.} \bibinfo{year}{2021}\natexlab{}.
\newblock \showarticletitle{The inductive bias of in-context learning: Rethinking pretraining example design}.
\newblock \bibinfo{journal}{\emph{arXiv preprint arXiv:2110.04541}} (\bibinfo{year}{2021}).
\newblock


\bibitem[Li et~al\mbox{.}(2021)]%
        {li2021adsgnn}
\bibfield{author}{\bibinfo{person}{Chaozhuo Li}, \bibinfo{person}{Bochen Pang}, \bibinfo{person}{Yuming Liu}, \bibinfo{person}{Hao Sun}, \bibinfo{person}{Zheng Liu}, \bibinfo{person}{Xing Xie}, \bibinfo{person}{Tianqi Yang}, \bibinfo{person}{Yanling Cui}, \bibinfo{person}{Liangjie Zhang}, {and} \bibinfo{person}{Qi Zhang}.} \bibinfo{year}{2021}\natexlab{}.
\newblock \showarticletitle{Adsgnn: Behavior-graph augmented relevance modeling in sponsored search}. In \bibinfo{booktitle}{\emph{Proceedings of the 44th international ACM SIGIR conference on research and development in information retrieval}}. \bibinfo{pages}{223--232}.
\newblock


\bibitem[Liu et~al\mbox{.}(2022)]%
        {liu2022aspect}
\bibfield{author}{\bibinfo{person}{Qidong Liu}, \bibinfo{person}{Cheng Long}, \bibinfo{person}{Jie Zhang}, \bibinfo{person}{Mingliang Xu}, {and} \bibinfo{person}{Dacheng Tao}.} \bibinfo{year}{2022}\natexlab{}.
\newblock \showarticletitle{Aspect-aware graph attention network for heterogeneous information networks}.
\newblock \bibinfo{journal}{\emph{IEEE Transactions on Neural Networks and Learning Systems}} (\bibinfo{year}{2022}).
\newblock


\bibitem[Liu(2019a)]%
        {liu2019fine}
\bibfield{author}{\bibinfo{person}{Yang Liu}.} \bibinfo{year}{2019}\natexlab{a}.
\newblock \showarticletitle{Fine-tune BERT for extractive summarization}.
\newblock \bibinfo{journal}{\emph{arXiv preprint arXiv:1903.10318}} (\bibinfo{year}{2019}).
\newblock


\bibitem[Liu(2019b)]%
        {liu2019roberta}
\bibfield{author}{\bibinfo{person}{Yinhan Liu}.} \bibinfo{year}{2019}\natexlab{b}.
\newblock \showarticletitle{Roberta: A robustly optimized bert pretraining approach}.
\newblock \bibinfo{journal}{\emph{arXiv preprint arXiv:1907.11692}} (\bibinfo{year}{2019}).
\newblock


\bibitem[Loshchilov(2017)]%
        {loshchilov2017decoupled}
\bibfield{author}{\bibinfo{person}{I Loshchilov}.} \bibinfo{year}{2017}\natexlab{}.
\newblock \showarticletitle{Decoupled weight decay regularization}.
\newblock \bibinfo{journal}{\emph{arXiv preprint arXiv:1711.05101}} (\bibinfo{year}{2017}).
\newblock


\bibitem[Meng et~al\mbox{.}(2022)]%
        {meng2022topic}
\bibfield{author}{\bibinfo{person}{Yu Meng}, \bibinfo{person}{Yunyi Zhang}, \bibinfo{person}{Jiaxin Huang}, \bibinfo{person}{Yu Zhang}, {and} \bibinfo{person}{Jiawei Han}.} \bibinfo{year}{2022}\natexlab{}.
\newblock \showarticletitle{Topic discovery via latent space clustering of pretrained language model representations}. In \bibinfo{booktitle}{\emph{Proceedings of the ACM web conference 2022}}. \bibinfo{pages}{3143--3152}.
\newblock


\bibitem[Mikolov et~al\mbox{.}(2013)]%
        {mikolov2013distributed}
\bibfield{author}{\bibinfo{person}{Tomas Mikolov}, \bibinfo{person}{Ilya Sutskever}, \bibinfo{person}{Kai Chen}, \bibinfo{person}{Greg~S Corrado}, {and} \bibinfo{person}{Jeff Dean}.} \bibinfo{year}{2013}\natexlab{}.
\newblock \showarticletitle{Distributed representations of words and phrases and their compositionality}.
\newblock \bibinfo{journal}{\emph{Advances in neural information processing systems}}  \bibinfo{volume}{26} (\bibinfo{year}{2013}).
\newblock


\bibitem[OpenAI(2023)]%
        {openai2023gpt}
\bibfield{author}{\bibinfo{person}{R OpenAI}.} \bibinfo{year}{2023}\natexlab{}.
\newblock \showarticletitle{Gpt-4 technical report. arxiv 2303.08774}.
\newblock \bibinfo{journal}{\emph{View in Article}} \bibinfo{volume}{2}, \bibinfo{number}{5} (\bibinfo{year}{2023}).
\newblock


\bibitem[Pennington et~al\mbox{.}(2014)]%
        {pennington2014glove}
\bibfield{author}{\bibinfo{person}{Jeffrey Pennington}, \bibinfo{person}{Richard Socher}, {and} \bibinfo{person}{Christopher~D Manning}.} \bibinfo{year}{2014}\natexlab{}.
\newblock \showarticletitle{Glove: Global vectors for word representation}. In \bibinfo{booktitle}{\emph{Proceedings of the 2014 conference on empirical methods in natural language processing (EMNLP)}}. \bibinfo{pages}{1532--1543}.
\newblock


\bibitem[Radford et~al\mbox{.}(2019)]%
        {radford2019language}
\bibfield{author}{\bibinfo{person}{Alec Radford}, \bibinfo{person}{Jeff Wu}, \bibinfo{person}{Rewon Child}, \bibinfo{person}{David Luan}, \bibinfo{person}{Dario Amodei}, {and} \bibinfo{person}{Ilya Sutskever}.} \bibinfo{year}{2019}\natexlab{}.
\newblock \showarticletitle{Language Models are Unsupervised Multitask Learners}.
\newblock  (\bibinfo{year}{2019}).
\newblock


\bibitem[Raffel et~al\mbox{.}(2020)]%
        {2020t5}
\bibfield{author}{\bibinfo{person}{Colin Raffel}, \bibinfo{person}{Noam Shazeer}, \bibinfo{person}{Adam Roberts}, \bibinfo{person}{Katherine Lee}, \bibinfo{person}{Sharan Narang}, \bibinfo{person}{Michael Matena}, \bibinfo{person}{Yanqi Zhou}, \bibinfo{person}{Wei Li}, {and} \bibinfo{person}{Peter~J. Liu}.} \bibinfo{year}{2020}\natexlab{}.
\newblock \showarticletitle{Exploring the Limits of Transfer Learning with a Unified Text-to-Text Transformer}.
\newblock \bibinfo{journal}{\emph{Journal of Machine Learning Research}} \bibinfo{volume}{21}, \bibinfo{number}{140} (\bibinfo{year}{2020}), \bibinfo{pages}{1--67}.
\newblock
\urldef\tempurl%
\url{http://jmlr.org/papers/v21/20-074.html}
\showURL{%
\tempurl}


\bibitem[Sanh(2019)]%
        {sanh2019distilbert}
\bibfield{author}{\bibinfo{person}{V Sanh}.} \bibinfo{year}{2019}\natexlab{}.
\newblock \showarticletitle{DistilBERT, A Distilled Version of BERT: Smaller, Faster, Cheaper and Lighter}.
\newblock \bibinfo{journal}{\emph{arXiv preprint arXiv:1910.01108}} (\bibinfo{year}{2019}).
\newblock


\bibitem[Shi et~al\mbox{.}(2019)]%
        {shi2019discovering}
\bibfield{author}{\bibinfo{person}{Yu Shi}, \bibinfo{person}{Jiaming Shen}, \bibinfo{person}{Yuchen Li}, \bibinfo{person}{Naijing Zhang}, \bibinfo{person}{Xinwei He}, \bibinfo{person}{Zhengzhi Lou}, \bibinfo{person}{Qi Zhu}, \bibinfo{person}{Matthew Walker}, \bibinfo{person}{Myunghwan Kim}, {and} \bibinfo{person}{Jiawei Han}.} \bibinfo{year}{2019}\natexlab{}.
\newblock \showarticletitle{Discovering hypernymy in text-rich heterogeneous information network by exploiting context granularity}. In \bibinfo{booktitle}{\emph{Proceedings of the 28th ACM International Conference on Information and Knowledge Management}}. \bibinfo{pages}{599--608}.
\newblock


\bibitem[Sun et~al\mbox{.}(2011)]%
        {sun2011pathsim}
\bibfield{author}{\bibinfo{person}{Yizhou Sun}, \bibinfo{person}{Jiawei Han}, \bibinfo{person}{Xifeng Yan}, \bibinfo{person}{Philip~S Yu}, {and} \bibinfo{person}{Tianyi Wu}.} \bibinfo{year}{2011}\natexlab{}.
\newblock \showarticletitle{Pathsim: Meta path-based top-k similarity search in heterogeneous information networks}.
\newblock \bibinfo{journal}{\emph{Proceedings of the VLDB Endowment}} \bibinfo{volume}{4}, \bibinfo{number}{11} (\bibinfo{year}{2011}), \bibinfo{pages}{992--1003}.
\newblock


\bibitem[Tang et~al\mbox{.}(2024)]%
        {tang2024graphgpt}
\bibfield{author}{\bibinfo{person}{Jiabin Tang}, \bibinfo{person}{Yuhao Yang}, \bibinfo{person}{Wei Wei}, \bibinfo{person}{Lei Shi}, \bibinfo{person}{Lixin Su}, \bibinfo{person}{Suqi Cheng}, \bibinfo{person}{Dawei Yin}, {and} \bibinfo{person}{Chao Huang}.} \bibinfo{year}{2024}\natexlab{}.
\newblock \showarticletitle{Graphgpt: Graph instruction tuning for large language models}. In \bibinfo{booktitle}{\emph{Proceedings of the 47th International ACM SIGIR Conference on Research and Development in Information Retrieval}}. \bibinfo{pages}{491--500}.
\newblock


\bibitem[Tang et~al\mbox{.}(2008)]%
        {tang2008arnetminer}
\bibfield{author}{\bibinfo{person}{Jie Tang}, \bibinfo{person}{Jing Zhang}, \bibinfo{person}{Limin Yao}, \bibinfo{person}{Juanzi Li}, \bibinfo{person}{Li Zhang}, {and} \bibinfo{person}{Zhong Su}.} \bibinfo{year}{2008}\natexlab{}.
\newblock \showarticletitle{Arnetminer: extraction and mining of academic social networks}. In \bibinfo{booktitle}{\emph{Proceedings of the 14th ACM SIGKDD international conference on Knowledge discovery and data mining}}. \bibinfo{pages}{990--998}.
\newblock


\bibitem[Trouillon et~al\mbox{.}(2016)]%
        {trouillon2016complex}
\bibfield{author}{\bibinfo{person}{Th{\'e}o Trouillon}, \bibinfo{person}{Johannes Welbl}, \bibinfo{person}{Sebastian Riedel}, \bibinfo{person}{{\'E}ric Gaussier}, {and} \bibinfo{person}{Guillaume Bouchard}.} \bibinfo{year}{2016}\natexlab{}.
\newblock \showarticletitle{Complex embeddings for simple link prediction}. In \bibinfo{booktitle}{\emph{International conference on machine learning}}. PMLR, \bibinfo{pages}{2071--2080}.
\newblock


\bibitem[Wang et~al\mbox{.}(2019)]%
        {wang2019heterogeneous}
\bibfield{author}{\bibinfo{person}{Xiao Wang}, \bibinfo{person}{Houye Ji}, \bibinfo{person}{Chuan Shi}, \bibinfo{person}{Bai Wang}, \bibinfo{person}{Yanfang Ye}, \bibinfo{person}{Peng Cui}, {and} \bibinfo{person}{Philip~S Yu}.} \bibinfo{year}{2019}\natexlab{}.
\newblock \showarticletitle{Heterogeneous graph attention network}. In \bibinfo{booktitle}{\emph{The world wide web conference}}. \bibinfo{pages}{2022--2032}.
\newblock


\bibitem[Xun et~al\mbox{.}(2020)]%
        {xun2020correlation}
\bibfield{author}{\bibinfo{person}{Guangxu Xun}, \bibinfo{person}{Kishlay Jha}, \bibinfo{person}{Jianhui Sun}, {and} \bibinfo{person}{Aidong Zhang}.} \bibinfo{year}{2020}\natexlab{}.
\newblock \showarticletitle{Correlation networks for extreme multi-label text classification}. In \bibinfo{booktitle}{\emph{Proceedings of the 26th ACM SIGKDD International Conference on Knowledge Discovery \& Data Mining}}. \bibinfo{pages}{1074--1082}.
\newblock


\bibitem[Yang et~al\mbox{.}(2021b)]%
        {yang2021graphformers}
\bibfield{author}{\bibinfo{person}{Junhan Yang}, \bibinfo{person}{Zheng Liu}, \bibinfo{person}{Shitao Xiao}, \bibinfo{person}{Chaozhuo Li}, \bibinfo{person}{Defu Lian}, \bibinfo{person}{Sanjay Agrawal}, \bibinfo{person}{Amit Singh}, \bibinfo{person}{Guangzhong Sun}, {and} \bibinfo{person}{Xing Xie}.} \bibinfo{year}{2021}\natexlab{b}.
\newblock \showarticletitle{Graphformers: Gnn-nested transformers for representation learning on textual graph}.
\newblock \bibinfo{journal}{\emph{Advances in Neural Information Processing Systems}}  \bibinfo{volume}{34} (\bibinfo{year}{2021}), \bibinfo{pages}{28798--28810}.
\newblock


\bibitem[Yang et~al\mbox{.}(2021a)]%
        {yang2021interpretable}
\bibfield{author}{\bibinfo{person}{Yaming Yang}, \bibinfo{person}{Ziyu Guan}, \bibinfo{person}{Jianxin Li}, \bibinfo{person}{Wei Zhao}, \bibinfo{person}{Jiangtao Cui}, {and} \bibinfo{person}{Quan Wang}.} \bibinfo{year}{2021}\natexlab{a}.
\newblock \showarticletitle{Interpretable and efficient heterogeneous graph convolutional network}.
\newblock \bibinfo{journal}{\emph{IEEE Transactions on Knowledge and Data Engineering}} \bibinfo{volume}{35}, \bibinfo{number}{2} (\bibinfo{year}{2021}), \bibinfo{pages}{1637--1650}.
\newblock


\bibitem[Yang et~al\mbox{.}(2022)]%
        {yang2022self}
\bibfield{author}{\bibinfo{person}{Yaming Yang}, \bibinfo{person}{Ziyu Guan}, \bibinfo{person}{Zhe Wang}, \bibinfo{person}{Wei Zhao}, \bibinfo{person}{Cai Xu}, \bibinfo{person}{Weigang Lu}, {and} \bibinfo{person}{Jianbin Huang}.} \bibinfo{year}{2022}\natexlab{}.
\newblock \showarticletitle{Self-supervised heterogeneous graph pre-training based on structural clustering}.
\newblock \bibinfo{journal}{\emph{Advances in Neural Information Processing Systems}}  \bibinfo{volume}{35} (\bibinfo{year}{2022}), \bibinfo{pages}{16962--16974}.
\newblock


\bibitem[Yasunaga et~al\mbox{.}(2022)]%
        {yasunaga2022linkbert}
\bibfield{author}{\bibinfo{person}{Michihiro Yasunaga}, \bibinfo{person}{Jure Leskovec}, {and} \bibinfo{person}{Percy Liang}.} \bibinfo{year}{2022}\natexlab{}.
\newblock \showarticletitle{Linkbert: Pretraining language models with document links}.
\newblock \bibinfo{journal}{\emph{arXiv preprint arXiv:2203.15827}} (\bibinfo{year}{2022}).
\newblock


\bibitem[Ye et~al\mbox{.}(2024)]%
        {ye2024language}
\bibfield{author}{\bibinfo{person}{Ruosong Ye}, \bibinfo{person}{Caiqi Zhang}, \bibinfo{person}{Runhui Wang}, \bibinfo{person}{Shuyuan Xu}, {and} \bibinfo{person}{Yongfeng Zhang}.} \bibinfo{year}{2024}\natexlab{}.
\newblock \showarticletitle{Language is all a graph needs}. In \bibinfo{booktitle}{\emph{Findings of the Association for Computational Linguistics: EACL 2024}}. \bibinfo{pages}{1955--1973}.
\newblock


\bibitem[Yu et~al\mbox{.}(2021)]%
        {yu2021gcn}
\bibfield{author}{\bibinfo{person}{Zhizhi Yu}, \bibinfo{person}{Di Jin}, \bibinfo{person}{Ziyang Liu}, \bibinfo{person}{Dongxiao He}, \bibinfo{person}{Xiao Wang}, \bibinfo{person}{Hanghang Tong}, {and} \bibinfo{person}{Jiawei Han}.} \bibinfo{year}{2021}\natexlab{}.
\newblock \showarticletitle{AS-GCN: Adaptive semantic architecture of graph convolutional networks for text-rich networks}. In \bibinfo{booktitle}{\emph{2021 IEEE International Conference on Data Mining (ICDM)}}. IEEE, \bibinfo{pages}{837--846}.
\newblock


\bibitem[Yu et~al\mbox{.}(2023a)]%
        {yu2023embedding}
\bibfield{author}{\bibinfo{person}{Zhizhi Yu}, \bibinfo{person}{Di Jin}, \bibinfo{person}{Ziyang Liu}, \bibinfo{person}{Dongxiao He}, \bibinfo{person}{Xiao Wang}, \bibinfo{person}{Hanghang Tong}, {and} \bibinfo{person}{Jiawei Han}.} \bibinfo{year}{2023}\natexlab{a}.
\newblock \showarticletitle{Embedding text-rich graph neural networks with sequence and topical semantic structures}.
\newblock \bibinfo{journal}{\emph{Knowledge and Information Systems}} \bibinfo{volume}{65}, \bibinfo{number}{2} (\bibinfo{year}{2023}), \bibinfo{pages}{613--640}.
\newblock


\bibitem[Yu et~al\mbox{.}(2023b)]%
        {yu2023teko}
\bibfield{author}{\bibinfo{person}{Zhizhi Yu}, \bibinfo{person}{Di Jin}, \bibinfo{person}{Jianguo Wei}, \bibinfo{person}{Yawen Li}, \bibinfo{person}{Ziyang Liu}, \bibinfo{person}{Yue Shang}, \bibinfo{person}{Jiawei Han}, {and} \bibinfo{person}{Lingfei Wu}.} \bibinfo{year}{2023}\natexlab{b}.
\newblock \showarticletitle{TeKo: Text-Rich Graph Neural Networks With External Knowledge}.
\newblock \bibinfo{journal}{\emph{IEEE Transactions on Neural Networks and Learning Systems}} (\bibinfo{year}{2023}).
\newblock


\bibitem[Zhang et~al\mbox{.}(2019b)]%
        {zhang2019shne}
\bibfield{author}{\bibinfo{person}{Chuxu Zhang}, \bibinfo{person}{Ananthram Swami}, {and} \bibinfo{person}{Nitesh~V Chawla}.} \bibinfo{year}{2019}\natexlab{b}.
\newblock \showarticletitle{Shne: Representation learning for semantic-associated heterogeneous networks}. In \bibinfo{booktitle}{\emph{Proceedings of the twelfth ACM international conference on web search and data mining}}. \bibinfo{pages}{690--698}.
\newblock


\bibitem[Zhang et~al\mbox{.}(2019a)]%
        {zhang2019oag}
\bibfield{author}{\bibinfo{person}{Fanjin Zhang}, \bibinfo{person}{Xiao Liu}, \bibinfo{person}{Jie Tang}, \bibinfo{person}{Yuxiao Dong}, \bibinfo{person}{Peiran Yao}, \bibinfo{person}{Jie Zhang}, \bibinfo{person}{Xiaotao Gu}, \bibinfo{person}{Yan Wang}, \bibinfo{person}{Bin Shao}, \bibinfo{person}{Rui Li}, {et~al\mbox{.}}} \bibinfo{year}{2019}\natexlab{a}.
\newblock \showarticletitle{OAG: Toward linking large-scale heterogeneous entity graphs}. In \bibinfo{booktitle}{\emph{Proceedings of the 25th ACM SIGKDD international conference on knowledge discovery \& data mining}}. \bibinfo{pages}{2585--2595}.
\newblock


\bibitem[Zhu et~al\mbox{.}(2021)]%
        {zhu2021textgnn}
\bibfield{author}{\bibinfo{person}{Jason Zhu}, \bibinfo{person}{Yanling Cui}, \bibinfo{person}{Yuming Liu}, \bibinfo{person}{Hao Sun}, \bibinfo{person}{Xue Li}, \bibinfo{person}{Markus Pelger}, \bibinfo{person}{Tianqi Yang}, \bibinfo{person}{Liangjie Zhang}, \bibinfo{person}{Ruofei Zhang}, {and} \bibinfo{person}{Huasha Zhao}.} \bibinfo{year}{2021}\natexlab{}.
\newblock \showarticletitle{Textgnn: Improving text encoder via graph neural network in sponsored search}. In \bibinfo{booktitle}{\emph{Proceedings of the Web Conference 2021}}. \bibinfo{pages}{2848--2857}.
\newblock


\bibitem[Zou et~al\mbox{.}(2023)]%
        {zou2023pretraining}
\bibfield{author}{\bibinfo{person}{Tao Zou}, \bibinfo{person}{Le Yu}, \bibinfo{person}{Yifei Huang}, \bibinfo{person}{Leilei Sun}, {and} \bibinfo{person}{Bowen Du}.} \bibinfo{year}{2023}\natexlab{}.
\newblock \showarticletitle{Pretraining language models with text-attributed heterogeneous graphs}.
\newblock \bibinfo{journal}{\emph{arXiv preprint arXiv:2310.12580}} (\bibinfo{year}{2023}).
\newblock


\end{thebibliography}

\appendix
\section{Dataset Description}
\label{sec:dataset}
The details of two HTRN benchmark datasets used in this paper are described as follows. 

\begin{itemize}[leftmargin=*]
\item \textbf{DBLP~\cite{tang2008arnetminer}.} The dataset is extracted from four fields of DBLP bibliography\footnote{\url{https://dblp.org}}. The four fields are: mathematical optimization, pattern recognition, computer vision and computer network. Based on the dataset, we construct an HTRN containing three types of nodes: 53,614 papers (P), 10,279 authors (A) and 3,524 venues (V), and three types of relations: 50,330 paper-paper (PP) edges, 47,573 paper-author (PA) edges and 51,642 paper-venue (PV) edges. We treat papers as text-rich nodes and extract the title and abstract parts as their textual information. Authors and venues are regarded as textless nodes with {\zhu only} name information. The target nodes, papers, are labeled according to the four fields.

\item \textbf{OAG~\cite{zhang2019oag}.} The dataset is a subgraph extracted from OAG based on five venues: IEEE Transactions on Biomedical Engineering, IEEE Transactions on Vehicular Technology, IEEE Journal of Solid State Circuits, INTERACT and SIGGRAPH. Based on the dataset, we construct an HTRN containing four types of nodes: 41,632 papers (P), 20,366 authors (A) and 20 fields (F), 1,790 institutions (I) and four types of relations: 55,794 paper-paper (PP) edges, 73,584 paper-author (PA) edges 87,972 paper-field (PF) edges and 22,869 author-institution (AI) edges. {\zhu Papers are treated as text-rich nodes, with titles and abstracts serving as their textual information. Authors, fields and institutions are considered textless nodes with {\zhu only} name information. The target nodes, papers, are labeled based on the five venues.}
\end{itemize}


\section{Reproducibility} 
\label{sec:rep}
{\wwwqyz To balance effectiveness and efficiency, and in line with previous work~\cite{zou2023pretraining,jin2023heterformer} to ensure a fair comparison, we adopt the widely used PLM Bert-base as the starting point for fine-tuning.}
Following the official configuration of BERT-base (110M parameters, \cite{kenton2019bert}), we limit the input length of the text to 512 tokens. We initialize the model with weights in BERT-base checkpoint released from Transformers tools\footnote{\url{https://huggingface.co/google-bert}}. To optimize HierPromptLM, we optimize the model with AdamW~\cite{loshchilov2017decoupled} with a learning rate searched in [1e-5, 5e-5]. We set the default masking ratio to 0.15 and the negative sampling ratio to 1. For fairness in comparison, we use a hidden dimension of 768 across all baselines and datasets, adopting hyperparameters recommended in the respective baseline papers. 

For the extension to PLM-based backbones, we adopt the same training objectives and follow the official configuration of T5-base\footnote{\url{https://huggingface.co/google-t5/t5-base}} (220M parameters, \cite{2020t5}) and GPT2-small\footnote{\url{https://huggingface.co/openai-community/gpt2}} (124M parameters, \cite{radford2019language}) as the backbone PLM to ensure a fair comparison.
All experiments are performed with one NVIDIA RTX A5000 GPU.  

\section{ADDITIONAL EXPERIMENTAL RESULTS.} 
\label{sec:add_exp}

\subsection{Link Prediction for Each Relation Type}
\label{sec:add_link}
we also conduct link prediction for each individual relation type on the DBLP dataset to further assess its effectiveness in capturing specific relations within HTRNs. As illustrated in Table~\ref{tab:add_link}, our HierPromptLM consistently shows superior performance in link prediction across all relation types, particularly excelling in the Paper-Paper relation with a significant {\zhu 28.38\% gain in ROC-AUC, 31.51\% enhancement in ROC-AUC, 21.40\% improvement in F1} over the second-best performance on the DBLP dataset. This highlights the efficacy of our pure PLM-based framework in learning heterogeneous edge representations within HTRNs.
Furthermore, the {\zhu M2V-Bert} model almost stands out among all baselines due to its effective utilization of meta-paths to capture heterogeneous topology and trace relations between nodes in HTRNs. Additionally, we observe that all baseline models struggle with the Paper-Paper relations. {\zhu This challenge arises because these relations often involve complex citations and references, requiring a deep understanding of the contextual connections between papers, which makes it difficult for the models to accurately capture these interactions.} 
{\zhu Moreover, although HTRN-based methods are specifically designed for HTRNs by modeling both textual and structural information, they may sometimes underperform compared to models that focus solely on text data or use a straightforward combination of textual and structural information for certain relation types.}
{\qy This reflects their inability to fully capture the heterogeneous graph structures, as well as the {\zhu critical} interactions between textual and structural information inherent within HTRNs}, which emphasizes the need for more advanced approaches such as our HierPromptLM for improving.



\subsection{Ablation Study on DBLP}
\label{sec:dblp_ablation}
We also conduct ablation study on the DBLP dataset, with results  shown in Table~\ref{tab:dblp_ablation}, {\zlwww which are consistent with the findings in} Section~\ref{sec:ablation}. Specifically, (1) {\zhu HierPromptLM significantly outperforms both w/o HGA-MLM and w/o HGA-NSP, highlighting the effectiveness of our two newly proposed HTRN-tailored tasks in capturing textual, heterogeneous structural information, and their  interactions within HTRNs. These results demonstrate the importance of designing specialized pretraining tasks for fine-tuning PLMs on HTRNs.} (2) The elimination of the meta-path-based graph token results in a marked performance decrease, reflecting the significance of meta-path-based subgraphs {\zhu for structural augmentation in enhancing representation learning on HTRNs.} (3) HierPromptLM {\zhu almost} outperforms w/o RelationToken, {\qy validating the importance of explicitly modeling heterogeneous relations with learnable embeddings for capturing edge-level heterogeneity within HTRNs.}

\begin{table*}[thb]
  \caption{Overall evaluation {\zhu on link prediction for each relation type}. Tabular results are in percent; the best results are highlighted in bold; the \underline{underlined} results indicate the second-best performance.}
  \label{tab:add_link}
  \vspace{-2mm}
  \setlength\tabcolsep{6pt}
  \centering
  \begin{tabular}{cccccccccc}
  \toprule
    \multirow{2}{*}{Methods} & \multicolumn{3}{c}{Paper-Paper} & \multicolumn{3}{c}{Paper-Author} & \multicolumn{3}{c}{Paper-Venue}  \\ 
  & ROC-AUC & PR-AUC & F1 & ROC-AUC & PR-AUC & F1 & ROC-AUC & PR-AUC & F1  \\ \hline
  {\zhu ComplEx+Bert} & 54.52 & 52.46 & 54.80 & 58.80 & 55.19 & 58.39 & 89.18 & 85.13 & 89.11 \\
  & ($\pm$0.03) & ($\pm$0.02) & ($\pm$0.14) & ($\pm$0.03) & ($\pm$0.02) & ($\pm$0.11) & ($\pm$0.02) & ($\pm$0.01) & ($\pm$0.02)\\
  {\zhu HIN2Vec+Bert} & 50.38 & 50.19 & 50.87 & 50.70 & 50.35 & 50.12 & 51.41 & 50.73 & 50.68 \\
   & ($\pm$0.00) & ($\pm$0.00) & ($\pm$0.01) & ($\pm$0.00) & ($\pm$0.00) & ($\pm$0.00) & ($\pm$0.00) & ($\pm$0.00) & ($\pm$0.00) \\
  {\zhu M2V+Bert} & \underline{64.66} & \underline{59.09} & \underline{68.17} & \underline{84.21} & \underline{77.74} & \underline{84.96} & 94.28 & 91.79 & 94.27 \\ 
  & ($\pm$0.93) & ($\pm$0.59) & ($\pm$1.94) & ($\pm$0.66) & ($\pm$0.50) & ($\pm$0.81) & ($\pm$0.04) & ($\pm$0.04) & ($\pm$0.04)\\ \hline
  {\zhu ie-HGCN+Bert} & 51.39 & 50.72 & 36.73 & 80.96 & 74.75 & 75.91 & \underline{95.66} & 93.01 & \underline{95.67} \\
  & ($\pm$1.50) & ($\pm$0.78) & ($\pm$29.37) & ($\pm$0.20) & ($\pm$9.42) & ($\pm$26.51) & ($\pm$2.31) & ($\pm$2.15) & ($\pm$2.54) \\ 
  {\zhu SHGP+Bert} & 51.61 & 50.84 & 55.65 & 80.54 & 74.53 & 74.53 & 91.64 & 89.17 & 89.04 \\ 
  & ($\pm$0.71) & ($\pm$0.38) & ($\pm$10.57) & ($\pm$6.93) & ($\pm$6.93) & ($\pm$5.51) & ($\pm$10.81) & ($\pm$10.12) & ($\pm$18.56)  \\  \hline
  Bert & 50.27 & 50.14 & 35.35 & 80.05 & 73.88 & 76.27 & 95.34 & 92.65 & 95.13 \\
   & ($\pm$0.29) & ($\pm$0.15) & ($\pm$31.09) & ($\pm$11.28) & ($\pm$8.95) & ($\pm$21.11) & ($\pm$4.42) & ($\pm$4.11) & ($\pm$5.32)   \\ 
  Albert & 51.76 & 50.91 & 53.41 & 81.86 & 75.33 & 79.95 & 94.87 & 92.20 & 94.75 \\ 
  & ($\pm$0.79) & ($\pm$0.41) & ($\pm$19.62) & ($\pm$9.65) & ($\pm$7.65) & ($\pm$16.46) & ($\pm$3.60) & ($\pm$3.33) & ($\pm$4.06) \\  \hline
  Heterformer & 60.68 & 56.55 & 62.17 & 79.38 & 77.36 & 84.28 & 93.28 & \underline{93.04} & 95.54 \\
  & ($\pm$0.30) & ($\pm$0.15) & ($\pm$1.65) & ($\pm$0.05) & ($\pm$0.03) & ($\pm$0.08) & ($\pm$0.01) & ($\pm$0.01) & ($\pm$0.01)  \\ 
  THLM & 52.13 & 51.16 & 41.81 & 84.05 & 77.21 & 84.83 & 95.32 & 92.70 & 95.35 \\ 
  & ($\pm$1.12) & ($\pm$0.62) & ($\pm$17.05) & ($\pm$3.40) & ($\pm$2.64) & ($\pm$4.47) & ($\pm$1.18) & ($\pm$1.04) & ($\pm$1.28)  \\  \hline
  {\zhu HierPromptLM} & \textbf{83.01} & \textbf{77.71} & \textbf{82.76} & \textbf{86.99} & \textbf{79.39} & \textbf{88.47} & \textbf{96.84} & \textbf{94.06} & \textbf{96.94} \\ 
  & ($\pm$0.00) & ($\pm$0.01) & ($\pm$0.01) & ($\pm$0.00) & ($\pm$0.00) & ($\pm$0.00) & ($\pm$0.00) & ($\pm$0.00) & ($\pm$0.00)  \\  \hline
  Improvement & \textbf{28.38\%} & \textbf{31.51\%} & \textbf{21.40\%} & \textbf{3.30\%} & \textbf{2.12\%} & \textbf{4.13\%} & \textbf{1.14\%} & \textbf{1.06\%} & \textbf{1.16\%} \\
  \bottomrule
  \end{tabular}
\end{table*}

\begin{table}[h]
  \caption{Ablation {\zhu study on DBLP}. Tabular results are in percent; the best results are highlighted in bold.}
  \label{tab:dblp_ablation}
  \vspace{-2mm}
  \setlength\tabcolsep{1.5pt}
  \centering
  \resizebox{0.48\textwidth}{!}{\begin{tabular}{cccccc}
  \toprule
    \multirow{2}{*}{Methods} & \multicolumn{2}{c}{Node classification} & \multicolumn{3}{c}{Link Prediction}  \\ 
   & Micro-F1 & Macro-F1 & ROC-AUC & PR-AUC & F1  \\ \hline

  w/o HGA-MLM & 54.62 & 17.66 & 50.00 & 50.00 & 33.33 \\ 
  & ($\pm$0.00) & ($\pm$0.00) & ($\pm$0.00) & ($\pm$0.00) & ($\pm$0.00)  \\
  w/o HGA-NSP & 87.48 & 84.57 & 77.68 & 73.06 & 75.66 \\ 
  & ($\pm$0.40) & ($\pm$0.51) & ($\pm$0.32) & ($\pm$0.79) & ($\pm$0.22)  \\
  w/o GraphToken & 87.45 & 84.76 & 80.15 & 75.56 & 78.74 \\ 
  & ($\pm$0.11) & ($\pm$0.15) & ($\pm$0.07) & ($\pm$0.02) & ($\pm$0.15) \\
  w/o RelationToken & 90.32 & \textbf{88.30} & 85.46 & 80.41 & 85.40  \\ 
  & ($\pm$0.08) & ($\pm$0.11) & ($\pm$0.01) & ($\pm$0.00) & ($\pm$0.01)  \\ \hline

  {\zhu HierPromptLM} & \textbf{90.36} & 88.15 & \textbf{85.56} & \textbf{80.44} & \textbf{85.56}  \\ 
  & ($\pm$0.07) & ($\pm$0.09) & ($\pm$0.04) & ($\pm$0.01) & ($\pm$0.05)  \\  
  \bottomrule
  \end{tabular}}
  \vspace{-10mm}
\end{table}

{\qywww \subsection{Evaluation of Training-free Setting on {\zhu DBLP}}
\label{sec:dblp_free}
We also evaluate the training-free variant of HierPromptLM on the DBLP dataset, with results displayed in Figure~\ref{fig:dblp_free}. These findings align with the observations discussed in  Section~\ref{sec:oag_free}. Particularly, HierPromptLM-free outperforms both Best Baselines and PLM-free across all metrics, showcasing the strong generalization capability of our model even without fine-tuning. {\qywww Additionally, HierPromptLM-free consistently demonstrates inferior performance compared to the fine-tuned version, HierPromptLM, highlighting the effectiveness of our proposed HTRN-tailored pretraining tasks for HTRN representation learning.}}

\begin{figure}[thb]
    \centering     \includegraphics[width=0.4\textwidth]{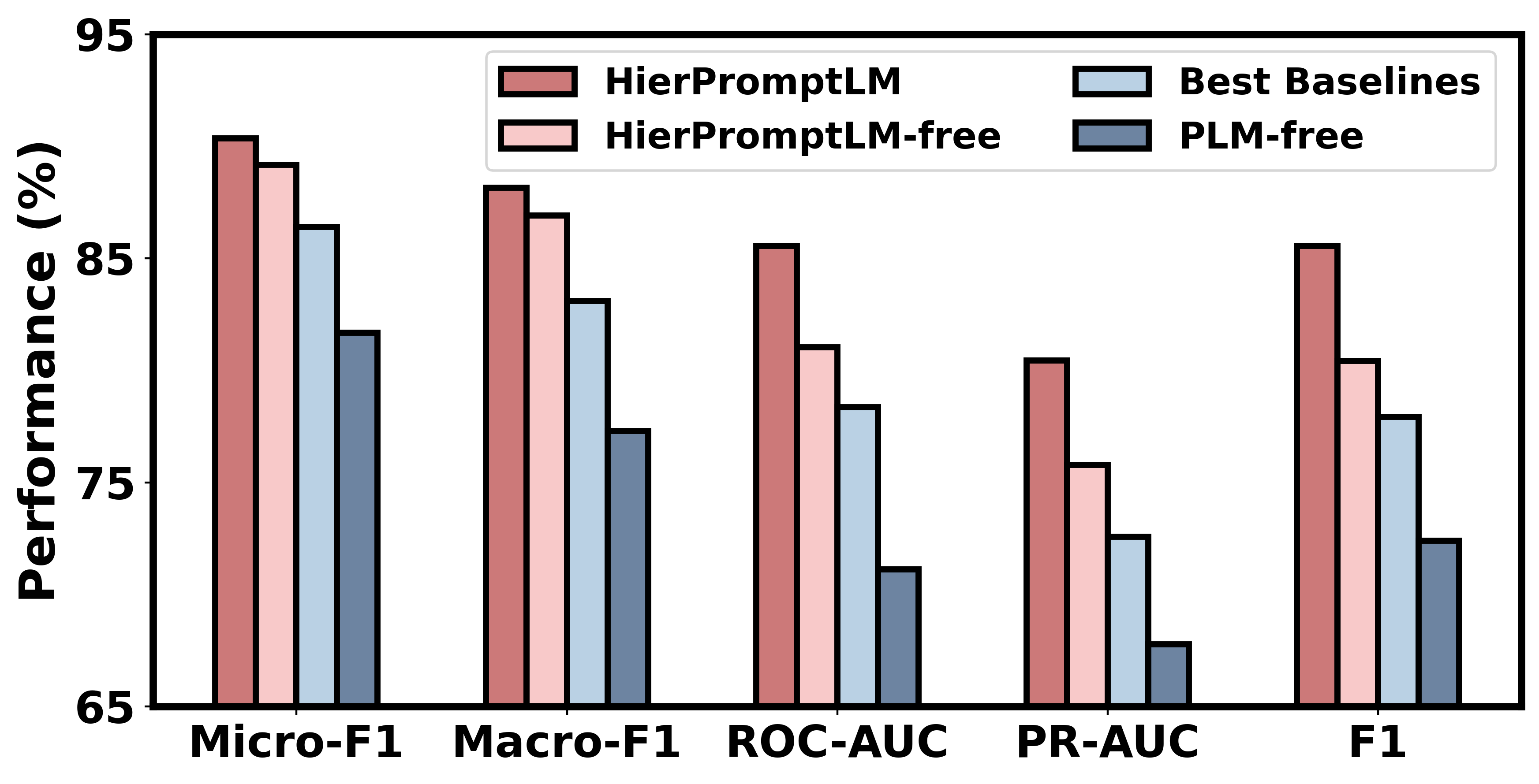}
    \vspace{-3mm}
    \caption{{\zhu Training-free extension on DBLP.}}
    \label{fig:dblp_free}
    \vspace{-4mm}
\end{figure}

\begin{figure}[thb]
    \centering     \includegraphics[width=0.32\textwidth]{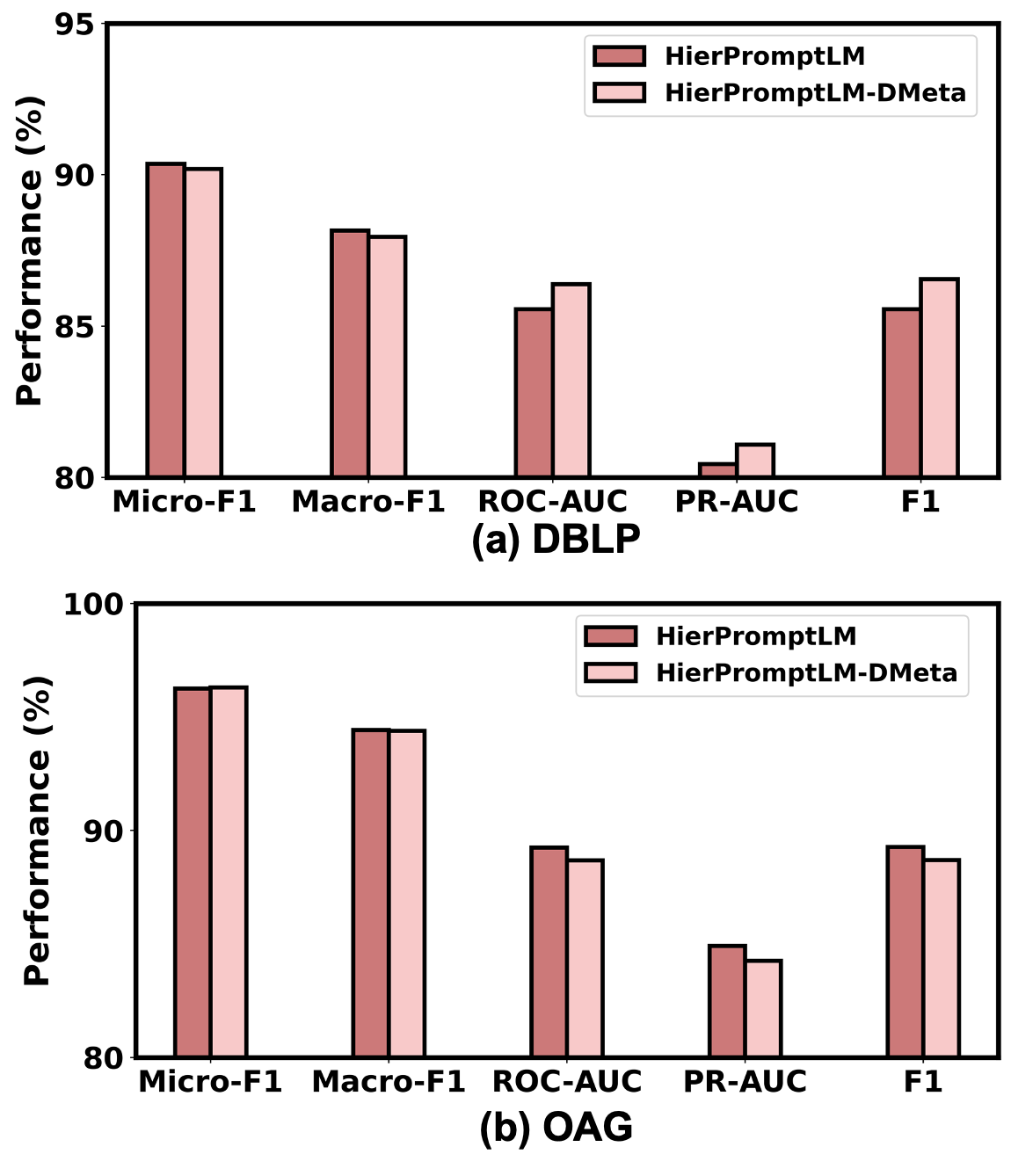}
    \vspace{-4mm}
    \caption{Impact of {\zhu pre-defined} meta-path.}
    \label{fig:meta}
    \vspace{-4mm}
\end{figure}


\subsection{Parameter Analysis}
\label{sec:add_para}
\noindent \textbf{Impact of Pre-defined Meta-paths.} 
To assess the impact of pre-defined meta-paths, we compare our HierPromptLM against a variant named \textbf{HierPromptLM-DMeta}, which uses simpler direct meta-paths. Specifically, for the DBLP dataset, HierPromptLM-DMeta {\zhu utilizes predefined meta-paths including paper-paper, paper-author, and paper-venue, whereas HierPromptLM expands on these with more complex paths such as paper-author-paper, paper-venue-paper, and author-paper-author.} Similarly, for the OAG dataset, HierPromptLM-DMeta employs paper-paper, paper-author, paper-field, and author-institution as {\zhu pre-defined} meta-paths, while HierPromptLM incorporates additional paths such as paper-author-paper, author-paper-author, and author-institution-author. The results for the DBLP and OAG datasets are shown in Figure~\ref{fig:meta}(a) and (b), respectively.

The results on the OAG dataset show that HierPromptLM generally outperforms HierPromptLM-DMeta. This is because HierPromptLM uses more complex meta-paths, such as paper-author-paper, paper-venue-paper and author-paper-author, which allows to capture richer heterogeneous structural information and better understand interactions within HTRNs.
However, on the DBLP dataset, HierPromptLM performs better than HierPromptLM-DMeta in node classification but tends to underperform in link prediction. The superior performance in node classification can be attributed to the model's ability to leverage the augmented structural information provided by more complex meta-paths, enhancing node representations and improving classification accuracy. In contrast, the decline in performance for link prediction could be due to the fact that {\qy on the DBLP dataset}, more direct relations (e.g., paper-paper or paper-author) are more relevant for predicting links. The inclusion of more complex meta-paths might introduce unnecessary noise or overcomplicate the representation for link prediction, where direct connections play a more critical role.

\begin{figure}[thb]
    \centering     \includegraphics[width=0.32\textwidth]{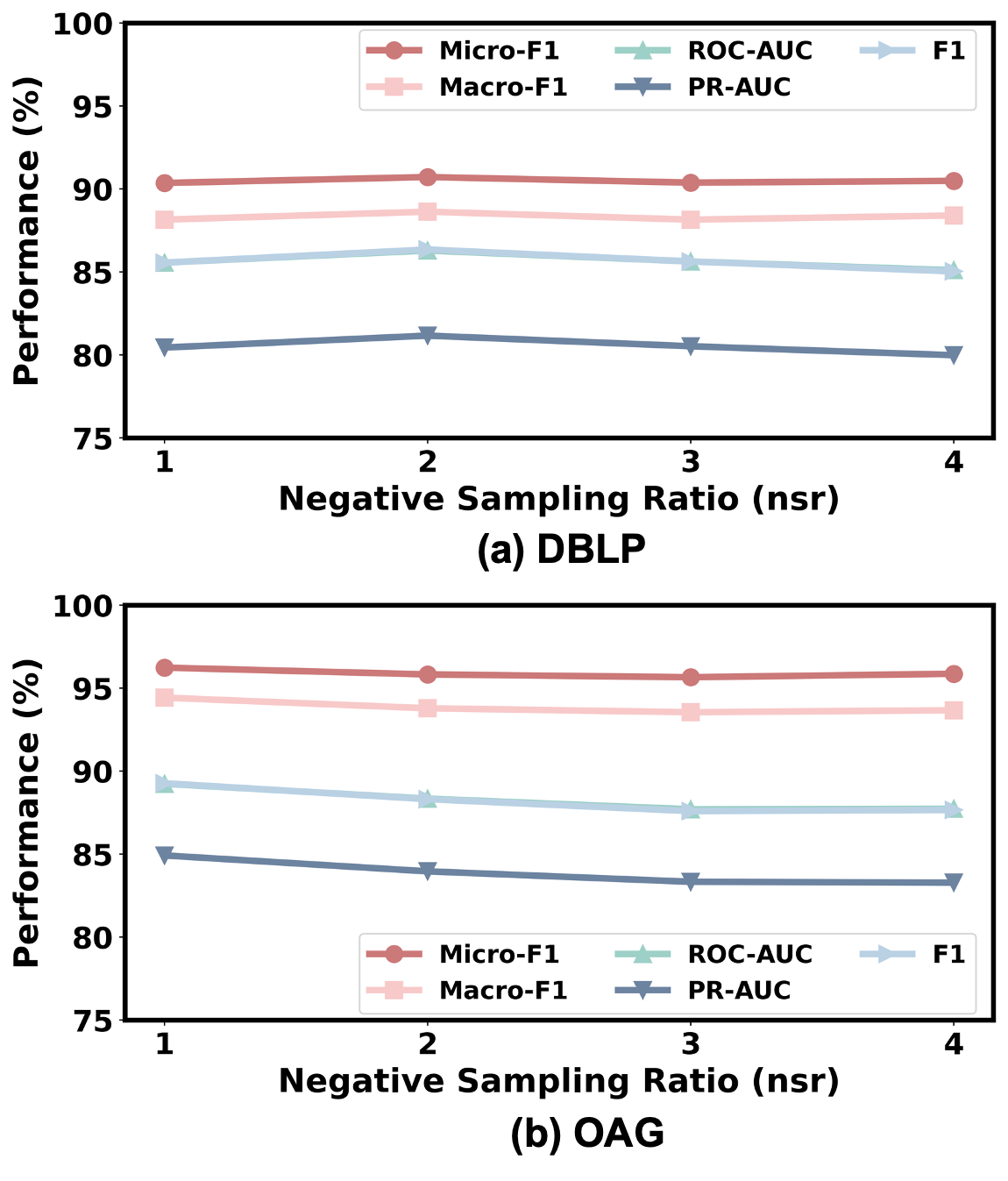}
    \vspace{-4mm}
    \caption{Impact of negative sampling ratio.}
    \label{fig:nsr}
    \vspace{-4mm}
\end{figure}


\begin{figure}[thb]
    \centering     \includegraphics[width=0.32\textwidth]{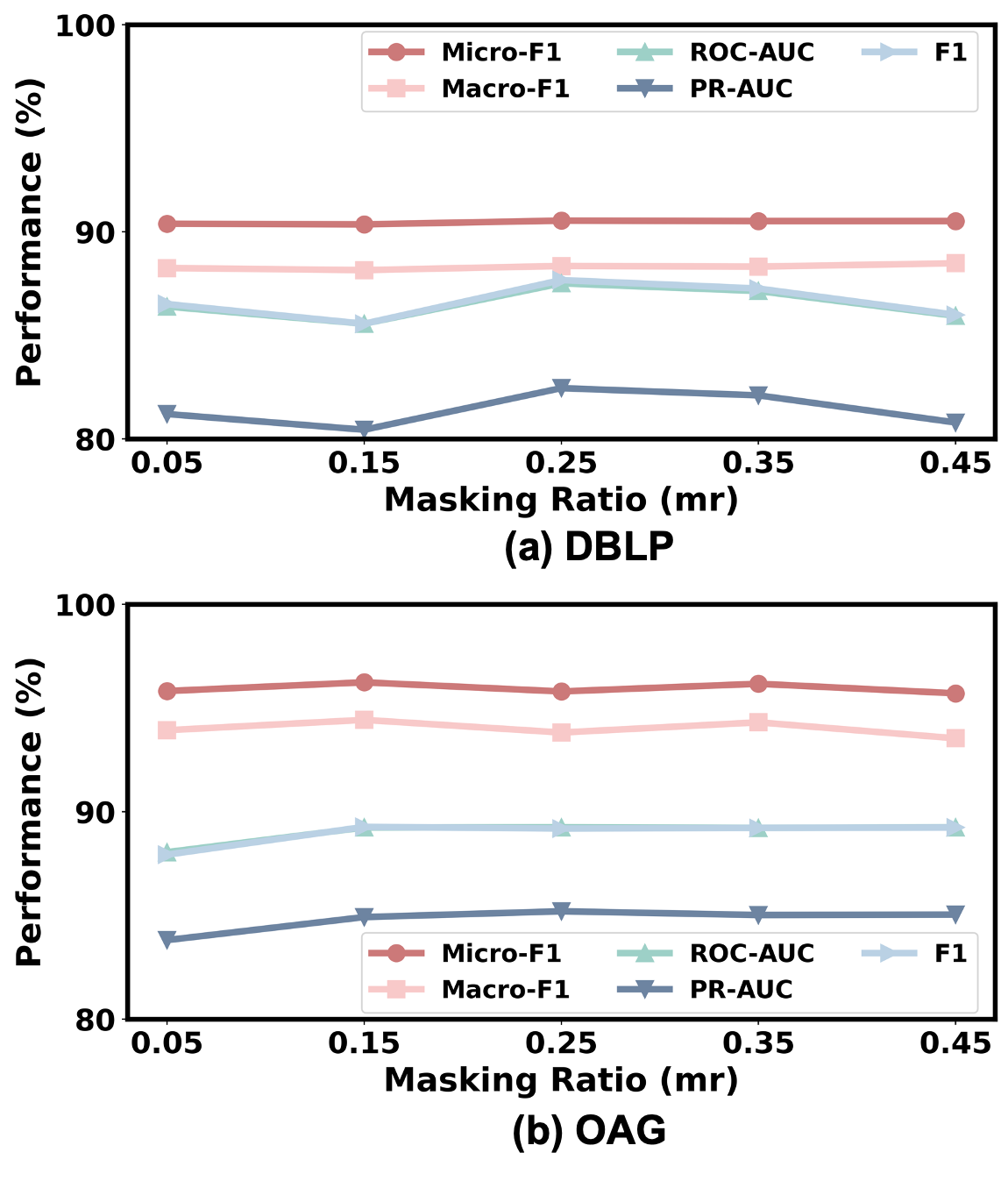}
    \vspace{-4mm}
    \caption{Impact of masking ratio.}
    \label{fig:mr}
    \vspace{-4mm}
\end{figure}


\noindent \textbf{Impact of Negative Sampling Ratio.} 
To validate the impact of negative sampling ratio $nsr$ in the HGA-NSP task, we vary $nsr$ in $\{1,2,3,4\}$, and the results on the DBLP and OAG dataset are shown in Figure~\ref{fig:nsr}(a) and (b), respectively. As observed, in most cases,  model performance generally declines as $nsr$ increases on the OAG dataset, while on the DBLP dataset, performance initially improves and then declines, with the best performance occurring at $nsr=1$ on the OAG dataset and $nsr=2$ on the DBLP dataset. This is because, at first, a larger negative sampling ratio provides more challenging negative samples, helping the model better distinguish between positive and negative examples. However, as $nsr$ continues to increase, the model might be overwhelmed by too many negative samples, causing it to focus excessively on distinguishing negatives rather than learning meaningful positive patterns. This imbalance can lead to performance degradation.


\noindent \textbf{Impact of Masking Ratio.} To evaluate the impact of masking ratio $mr$ in the HGA-MLM task, we vary $mr$ in $\{0.05,0.15,0.25,0.35,0.45\}$, and the results on {\zhu the DBLP and OAG datasets} are shown in Figure~\ref{fig:mr}(a) and (b), respectively. As observed on the DBLP dataset, performance initially improves and then declines, with the best performance occurring at $mr=0.25$, which differs from the optimal value of $mr=0.15$ in BERT's standard MLM task. This implies that when adapting BERT’s standard MLM task to HTRNs, the proposed HTRN-tailored HGA-MLM task requires a higher masking ratio to better capture both textual and heterogeneous structural information on the DBLP dataset, leading to more effective representation learning on HTRNs. Similarly, on the OAG dataset, performance initially improves and then declines, with the best performance occurring at $mr=0.15$ for node classification and $mr=0.25$ for link prediction. This indicates that different tasks within HTRNs may require different masking ratios for optimal performance, with node classification benefiting from a lower masking ratio, likely due to the importance of textual information, while link prediction benefits from a higher ratio, as it relies more heavily on the {\qy interactions} between structural and textual information inherent within HTRNs. {\qy These differences further emphasize the adaptability of the HGA-MLM task, demonstrating its ability to adjust to different aspects of HTRNs depending on the task requirements.}

\end{document}